\documentclass{article}

\PassOptionsToPackage{numbers, compress}{natbib}

\usepackage{graphicx}

\usepackage[final]{neurips_2025}

\usepackage{amsmath}

\usepackage[utf8]{inputenc} 
\usepackage[T1]{fontenc}    
\usepackage{hyperref}       
\usepackage{url}            
\usepackage{booktabs}       
\usepackage{amsfonts}       
\usepackage{nicefrac}       
\usepackage{microtype}      
\usepackage{xcolor}         

\usepackage{multirow}
\usepackage{tcolorbox}
\usepackage{color}
\usepackage{xcolor, ulem}
\newcommand{\rb}[1]{\textcolor{red}{\textbf{#1}}}
\newcommand{\bu}[1]{\textcolor{blue}{\uline{#1}}}

\title{TARFVAE: Efficient One-Step Generative Time Series Forecasting via TARFLOW based VAE}

\author{
  Jiawen Wei \\
  Meituan \\
  Beijing, China \\
  \texttt{weijiawen@meituan.com} \\
  \And
  Lan Jiang \\
  Meituan \\
  Beijing, China \\
  \texttt{jianglan09@meituan.com} \\
  \And
  Pengbo Wei \\
  Meituan \\
  Beijing, China \\
  \texttt{weipengbo@meituan.com} \\
  \And
  Ziwen Ye \\
  Meituan \\
  Beijing, China \\
  \texttt{yeziwen@meituan.com} \\
  \And
  Teng Song \\
  Meituan \\
  Beijing, China \\
  \texttt{songteng02@meituan.com} \\
  \And
  Chen Chen \\
  Meituan \\
  Beijing, China \\
  \texttt{chenchen11@meituan.com} \\
  \And
  Guangrui Ma\thanks{Corresponding author.} \\
  Meituan \\
  Beijing, China \\
  \texttt{magr@connect.ust.hk} \\ 
}

\begin{document}

\maketitle

\begin{abstract}
  Time series data is ubiquitous, with forecasting applications spanning from finance to healthcare. Beyond popular deterministic methods, generative models are gaining attention due to advancements in areas like image synthesis and video generation, as well as their inherent ability to provide probabilistic predictions. However, existing generative approaches mostly involve recurrent generative operations or repeated denoising steps, making the prediction laborious, particularly for long-term forecasting. Most of them only conduct experiments for relatively short-term forecasting, with limited comparison to deterministic methods in long-term forecasting, leaving their practical advantages unclear. This paper presents TARFVAE, a novel generative framework that combines the Transformer-based autoregressive flow (TARFLOW) and variational autoencoder (VAE) for efficient one-step generative time series forecasting. Inspired by the rethinking that complex architectures for extracting time series representations might not be necessary, we add a flow module, TARFLOW, to VAE to promote spontaneous learning of latent variables that benefit predictions. TARFLOW enhances VAE's posterior estimation by breaking the Gaussian assumption, thereby enabling a more informative latent space. TARFVAE uses only the forward process of TARFLOW, avoiding autoregressive inverse operations and thus ensuring fast generation. During generation, it samples from the prior latent space and directly generates full-horizon forecasts via the VAE decoder. With simple MLP modules, TARFVAE achieves superior performance over state-of-the-art deterministic and generative models across different forecast horizons on benchmark datasets while maintaining efficient prediction speed, demonstrating its effectiveness as an efficient and powerful solution for generative time series forecasting. Our code is available at https://github.com/Gavine77/TARFVAE.
\end{abstract}

\section{Introduction}

In various application fields such as transportation planning\cite{fang2022attention}, healthcare\cite{morid2023time} and inventory management\cite{qi2023practical}, long-term time series forecasting (LTSF) and uncertainty quantification are of vital importance. The former provides a solid basis for long-term decision-making by offering reliable prediction, while the latter delivers more comprehensive estimates, enabling robust decision-making through accounting for potential variability. Recently, deep learning has achieved remarkable success in these fields through capturing complex time series patterns\cite{nietime, liuitransformer, zhou2021informer, wu2021autoformer, zhou2022fedformer, ekambaram2023tsmixer} and generating reliable uncertainty estimates\cite{rasul2021autoregressive, tashiro2021csdi, salinas2020deepar, shen2023non}. 

In the realm of long-term forecasting, various strategies have been proposed to enhance predictive performance. Most efforts have focused on Transformer-based models due to their proven capability in capturing long-term dependencies via attention mechanisms\cite{zhou2021informer}. Numerous studies aim to reduce the computational complexity of attention\cite{zhou2021informer,wu2021autoformer,zhou2022fedformer} and improve information extraction, while others apply vanilla Transformers to mine intra-channel or inter-channel relationships\cite{nietime, liuitransformer}. However, some studies\cite{zeng2023transformers, sun2025simple} argue that sophisticated designs might not be necessary and suggest simple models could deliver results comparable to Transformer-based models. This has spurred exploration into Linear-based or MLP-based methods that aim at more efficient utilization of historical information during training to improve prediction performance\cite{ekambaram2023tsmixer, das2023longterm, hansofts}.

Current LTSF methods are predominantly deterministic, producing point estimates without systematically addressing uncertainty quantification. This oversight poses substantial risks to operational robustness, particularly given the intrinsic positive correlation between uncertainty magnitude and forecast horizon length. To achieve probabilistic forecasting, earlier models like DeepAR\cite{salinas2020deepar} employ recurrent neural networks (RNNs) to estimate parameters of prespecified distributions (e.g., Gaussian) for each timestep. However, such methods impose restrictive parametric assumptions that may not capture complex temporal dynamics. Subsequent advancements have explored generative frameworks, including variational autoencoders (VAEs)\cite{desai2021timevae} and diffusion models\cite{rasul2021autoregressive, shen2023non}, building upon their demonstrated success in adjacent domains\cite{dhariwal2021diffusion, harvey2022flexible, saharia2022photorealistic}. While diffusion models achieve state-of-the-art sample quality, their iterative denoising process incurs substantial computational overhead during inference. Recent attempts to accelerate generation\cite{tashiro2021csdi, shen2023non, li2022generative} remain fundamentally constrained by the multi-step generation paradigm. By contrast, conventional VAE\cite{kingma2013auto} enables one-step generation by decoding samples from the prior latent space. However, many existing hybrid methods\cite{cai2023hybrid,chen2022deep,wang2022learning} introduce complex recurrent structures to capture temporal dependencies and generate predictions autoregressively, which undermine this computational advantage. Due to computational constraints, these generative models have primarily been developed and evaluated in short-term settings, raising concerns about their long-term efficacy and accuracy. Recently, Zhang et al.\cite{zhang2024probts} systematically compared deterministic and probabilistic methods for long-term forecasting, whose results still highlight the need for models that can effectively address both point and probabilistic forecasting across diverse horizons.

Motivated by these considerations, we introduce TARFVAE, a novel VAE-based framework that strategically integrates two core objectives: (1) preserving the one-step generation capability of VAEs to ensure computationally efficient inference, and (2) promoting spontaneous learning of latent variables that guarantee robust probabilistic forecasting performance. TARFVAE achieves these by simply incorporating a Transformer-based autoregressive flow (TARFLOW)\cite{zhai2024normalizing}, which enables an enhanced posterior estimation. Note that the autoregressive inverse process of TARFLOW is not included so TARFVAE can perform one-step generation. Extensive experiments on eight real-world datasets demonstrate the superiority of TARFVAE over the existing state-of-the-art deterministic and generative baselines.

\section{Related Work}
\paragraph{Long-term Time Series Forecasting}

Transformer-based models have achieved remarkable success in LTSF through their attention mechanisms, which effectively capture long-term dependencies\cite{zhou2021informer,wu2021autoformer,zhou2022fedformer}. According to Qiu et al.\cite{qiu2025comprehensive}, these models can be categorized into channel-independent and channel-dependent approaches. Initially, they aggregate channel information via linear mapping and use attention to extract temporal patterns. However, the high computational complexity of attention mechanisms poses challenges in long-term forecasting. To address this, significant efforts have focused on enhancing attention efficiency\cite{zhou2021informer,wu2021autoformer,zhou2022fedformer, zhang2023crossformer}. Despite progress, channel-dependent strategies are more vulnerable to distributional shifts\cite{zeng2023transformers, han2024capacity}, prompting some researchers to explore channel-independent approaches, which have shown superior performance\cite{nietime,liuitransformer, zeng2023transformers}. For example, PatchTST\cite{nietime} splits historical data into patches and applies a standard Transformer to model their correlations. Meanwhile, other studies have leveraged simple Linear-based or MLP-based structures for time series modeling\cite{zeng2023transformers, das2023longterm, hansofts, han2024capacity}. For instance, SOFTS\cite{hansofts} introduces an MLP-based module to aggregate multiple channels and generate core representations for prediction.

\paragraph{Generative Models for LTSF} 
To boost model robustness and reliability, recent studies have shifted from estimating the median or mean of time series to capturing the complete time series distribution using generative models. Various generative models, such as VAEs\cite{kingma2013auto, cai2023hybrid}, normalizing flows\cite{rasul2020tempflow,papamakarios2021normalizing} and diffusion models\cite{rasul2021autoregressive}, have been applied in this field. Transformer-MAF\cite{rasul2020tempflow} is the pioneering work in applying normalzing flows: it builds upon the Transformer architecture and utilizes Masked Autoregressive Flows (MAF)\cite{papamakarios2017masked} for generating time series forecasts. Recently, diffusion models have gained prominence in LTSF because of their superior performance in adjacent domains\cite{dhariwal2021diffusion, harvey2022flexible, saharia2022photorealistic}. TimeGrad\cite{rasul2021autoregressive}, the first time series diffusion model, relied on an RNN to predict in an autoregressive manner, with the denoising process guided by hidden states generated by RNN. Instead of autoregressive decoding, CSDI\cite{tashiro2021csdi} and SSSD\cite{lopez2023diffusion} use self-supervised masking to guide the denoising process and make predictions. Methods like TimeDiff\cite{shen2023non} and mrDiff\cite{shen2024multi} utilize different networks to incorporate historical information as conditions and employ conditional diffusion models for time series generation. However, diffusion-based models, though effective, require multiple steps to generate time series from noise, thus slowing inference. Unlike diffusion models, VAEs can efficiently generate data by directly decoding latent representations. Previous studies\cite{desai2021timevae, cai2023hybrid,chen2022deep,wang2022learning} improve the learning of latent space by better representing historical sequences through complex recurrent or autoregressive operations, which compromise VAE's fast inference speed.

\section{Preliminaries}
\label{sec:preliminaries}

\subsection{Variational AutoEncoder}
A VAE\cite{kingma2013auto} is an unsupervised generative model that models the input data distribution by learning a probabilistic latent space as follows:
\begin{equation}
    p(x)=\int p(x,z)dz=\int p(x|z)p(z)dz=\int p(x|z)\frac{p(z)}{p(z|x)}p(z|x) dz=\mathbb{E}_{z\sim p(z|x)}[p(x|z)\frac{p(z)}{p(z|x)}] \label{eq:p_x}
\end{equation}
where $x$ is the input data and $z$ is its latent representations. The prior $p(z)$ is normally defined as a multivariate Gaussian distribution $N(0,I)$. The posterior $p(z|x)$ can be an arbitrary distribution and VAE approximates it as $q(z|x)=N(\mu(x),\sigma^2(x))$. Thus VAE learns the problem by maximizing the log-likelihood of (\ref{eq:p_x}) as
\begin{align}
\log p(x) &= \log\mathbb{E}_{z\sim q(z|x)}\left[p(x|z)\frac{p(z)}{q(z|x)}\right] \label{eq:origin}\\
 &\ge \mathbb{E}_{z\sim q(z|x)}\log\left[p(x|z)\frac{p(z)}{q(z|x)}\right] \label{eq:elbo}\\ 
&= \mathbb{E}_{z\sim q(z|x)}\left[\log p(x|z) - \log\frac{q(z|x)}{p(z)}\right] \\
&=\mathbb{E}_{z\sim q(z|x)}\left[\log p(x|z)\right] - KL(q(z|x)||p(z)) \label{eq:finalform}
\end{align}
where (\ref{eq:elbo}) is the evidence lower bound (ELBO) of (\ref{eq:origin}) accoring to Jensen's inequality. The first term in (\ref{eq:finalform}) leads to reconstruct  accurate $x$ from $z$, and the second term minimizes the difference between the prior and approximated posterior to prompt the latent space to capture meaningful data representations. This enables VAE to generate new samples via $z \sim p(z)$, $x \sim p(x|z)$.

VAE can be extended to conditional VAE by incorporating supervised labels $y$. The ELBO for $\log p(x|y)$ can be analogously derived as follows:
\begin{align}
\log p(x|y) &\ge \mathbb{E}_{z\sim q(z|x,y)}\left[\log p(x|z,y) - \log\frac{q(z|x,y)}{p(z|y)}\right] \label{eq:cvaeloss}\\ 
&=\mathbb{E}_{z\sim q(z|x,y)}\left[\log p(x|z,y)\right] - KL(q(z|x,y)||p(z|y))
\end{align}
where the prior, approximated posterior and reconstruction process are all conditioned on $y$.

\subsection{Normalizing Flow}
Normalizing flows\cite{papamakarios2021normalizing, tabak2013family} are invertible mappings $f:\mathcal{X}  \to \mathcal{Z}$ from $\mathbb{R}^D$ to $\mathbb{R}^D$. For $x \sim \mathcal{X}$ and $z \sim \mathcal{Z}$, the density $p_{\mathcal{X}}(x)$ can be expressed by 
\begin{equation}
    p_{\mathcal{X}}(x)=p_{\mathcal{Z}}(f(x))\left | \det (\frac{\partial f(x)}{\partial x}) \right |. 
\end{equation}

Normalizing flows have the property that the inverse $x=f^{-1}(z)$ is easy to evaluate and computing the Jacobian determinant takes $O(D)$ time. Therefore, once the model is trained, a generative model is automatically obtained via $z \sim \mathcal{Z}$, $x=f^{-1}(z)$. Real NVP\cite{dinh2016density} designs an affine coupling layer to meet the above two conditions:
\begin{equation}
\begin{cases}
z_1=x_1\\
z_2= (x_2 - t(x_1)) \odot \exp(s(x_1))
\end{cases}
\label{eq:realnvp}
\end{equation}
where $x$ can be randomly shuffled into two parts $[x_1,x_2]$, and $s(x_1)$ and $t(x_1)$ are learnable functions whose outputs match the shape of $x_2$. The inverse of (\ref{eq:realnvp}) is obvious and the Jacobian matrix of (\ref{eq:realnvp}) is a lower triangular matrix. The determinant of the Jacobian is the product of the diagonal elements, so the log-determinant is
\begin{equation}
    \log |\det \frac{\partial z}{\partial x}|=\sum_i s_i(x_1).
\label{eq:logdet}
\end{equation}

To achieve stronger nonlinearity, multiple coupling layers are composed together, creating a chain of mappings: $\mathcal{X} =\mathcal{Z}_0 \to\mathcal{Z}_1 \to\mathcal{Z}_2 \to... \to\mathcal{Z}_K=\mathcal{Z}$. The maximum likelihood estimation objective can then be written as
\begin{equation}
\log p_{\mathcal{X}}(x)=\log p_{\mathcal{Z}}(z)+\sum_{k=1}^K \log |\det \frac{\partial z_k}{\partial z_{k-1}}|.
\label{eq:flowloss}
\end{equation}

Recently, Zhai et al. proposed TARFLOW\cite{zhai2024normalizing} which could achieve more efficient nonlinear mappings. TARFLOW partitions $x$ into more parts as $[x_1,x_2,...,x_n]$ and performs transformation following a similar rule:
\begin{equation}
\begin{cases}
z_1=x_1\\
z_j= (x_j - t^j(x_{<j}))\odot \exp(s^j(x_{<j}))
\end{cases}
\label{eq:tarflow}
\end{equation}
where $j>1$ and $x_{<j}=[x_1,x_2,...,x_{j-1}]$. The $s(x_{<j})$ and $t(x_{<j})$ are efficiently implemented by causal Transformer. Notably, the inverse of (\ref{eq:tarflow}) becomes
\begin{equation}
\begin{cases}
x_1=z_1\\
x_j=z_j \odot\exp(-s^j(x_{<j})) + t^j(x_{<j})
\end{cases}
\end{equation}
where the inference of $x_j$ relies on $x_{<j}$, making the inverse process autoregressive.

\subsection{Time Series Forecasting}
 Time series forecasting uses historical time series $x \in \mathbb{R} ^{C\times L}$ to predict the future values $y \in \mathbb{R} ^{C\times H}$. Here, $C$ represents the number of variables or channels, $L$ is the length of the lookback window, and $H$ stands for the forecast horizon. $C>1$ corresponds to a multivariate time series; otherwise, it is termed univariate.

Deterministic methods perform point estimation via a function $f$ as $\hat y=f(x)$, whereas probabilistic methods model a distribution $\hat p(y|x)$ to approximate the true $p(y|x)$.

\section{TARFVAE}
\subsection{Overview}
\begin{figure*}[!htbp]
  \includegraphics[width=\textwidth]{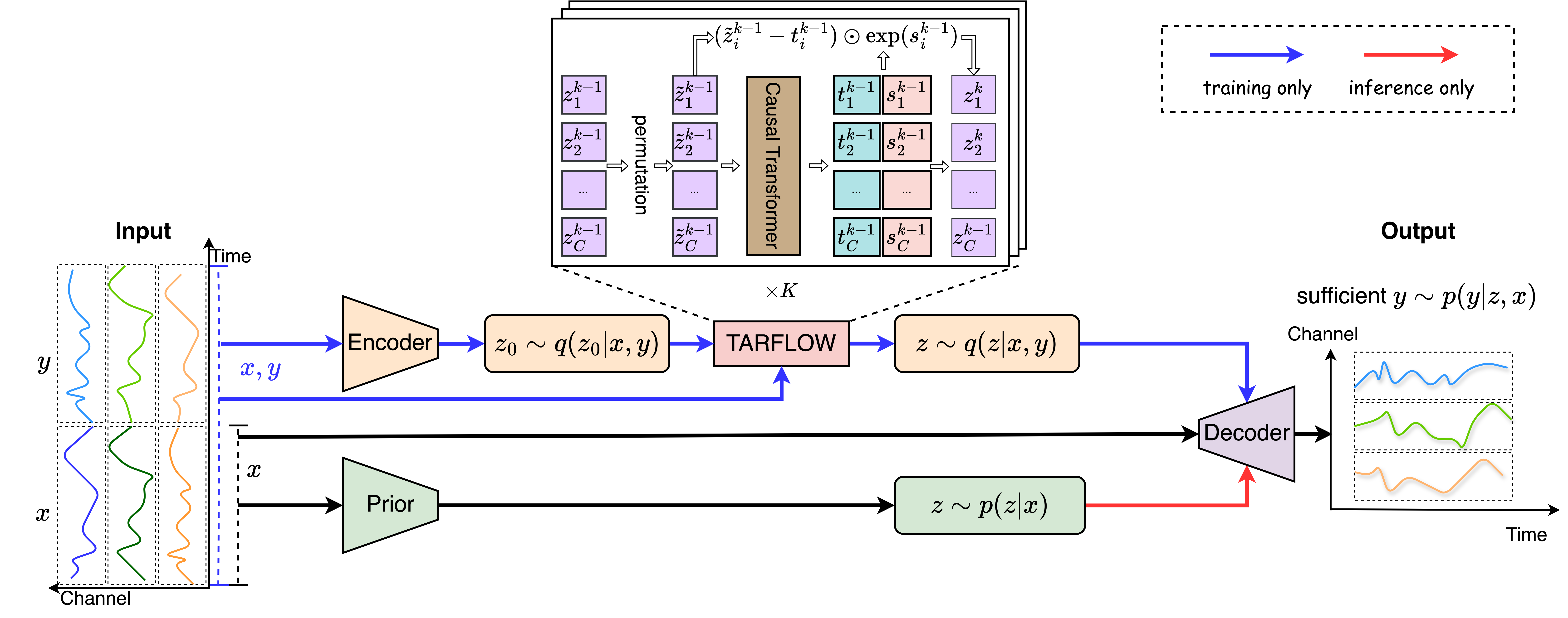}
  \caption{The overview of TARFVAE.}
  \label{fig:overview}
\end{figure*}

The architecture of TARFVAE, as illustrated in \textbf{Figure \ref{fig:overview}}, builds upon a conditional VAE where input history $x$ conditions the generation of target series $y$. The Gaussian posterior assumption in Section 3.1 fundamentally limits reconstruction accuracy in conventional VAEs, as Gaussian distributions constitute a tiny subset of all possible posterior distributions. To overcome this expressiveness limitation, we introduce TARFLOW to refine an initial Gaussian posterior $q(z_0|x,y)$ into a flexible complex distribution that closely matches the true posterior. Specifically, both $x$ and $y$ are utilized for training, whereas only $x$ is used for inference. 

\textbf{Instance normalization}. Following many state-of-the-art models\cite{nietime, liuitransformer,hansofts, qiu2024duet}, we first apply instance normalization\cite{kim2021reversible} to remove the local statistics of the input history to stabilize the base prediction, and then restore them to the model prediction. Notably, during training, both $x$ and $y$ are normalized based on $x$’s mean and variance without using $y$’s statistics. This ensures $x$’s normalization remains consistent between training and inference, regardless of $y$’s presence. 

\textbf{Series embedding}. Then we perform series embedding\cite{liuitransformer, hansofts} on the needed time series for the encoder, decoder, prior and flow modules independently. Each module employs a dedicated embedding layer to extract module-specific temporal patterns. This architectural isolation prevents cross-module interference while enabling targeted feature learning.

\textbf{One-step generation}. The prior and encoder modules parameterize two Gaussians, the prior $p(z|x)$ and initial posterior $q(z_0|x,y)$, respectively. During training, the flow module transforms $z_0 \sim q(z_0|x,y)$ to approximate $z$ from the true posterior, while inference directly samples $z \sim p(z|x)$. Finally, the decoder maps $z$ to $\hat y$ conditioned on $x$, enabling efficient one-step full-horizon generation.

The training and inference process of TARFVAE can be formulated as follows:

\textbf{Training}
\begin{align}
    & \mu_{prior},\log\sigma_{prior}^2=Prior(Emb_{1}(x))\\
    & \mu_{z_0},\log\sigma_{z_0}^2=Encoder(Emb_{2}([x,y]))\\
    & z_0\sim N(\mu_{z_0},\sigma_{z_0}^2)\\
    & z = TARFLOW(z_0,Emb_{3}([x,y]))\\
    & \hat y = Decoder(z,Emb_{4}(x)) 
\end{align}

\textbf{Inference}
\begin{align}
    & \mu_{prior},\log\sigma_{prior}^2=Prior(Emb_{1}(x))\\
    & z_{prior}\sim N(\mu_{prior},\sigma_{prior}^2)\\
    & \hat y = Decoder(z_{prior},Emb_{4}(x))
\end{align}

Below, we describe the TARFVAE implementation used in our work, while noting that alternative approaches could be adopted.

\subsection{MLP Foundation}
\begin{figure*}[!htbp]
  \centering
  \includegraphics[width=0.6\textwidth]{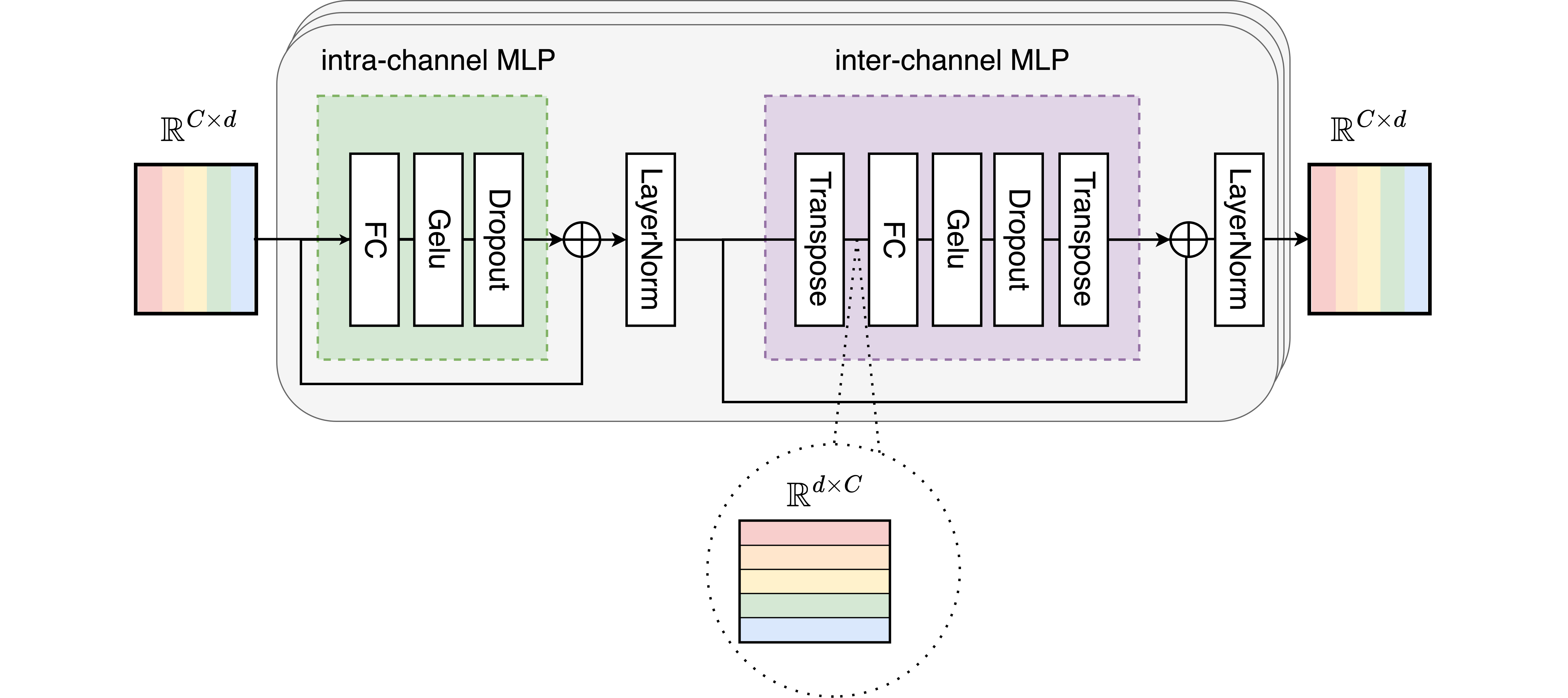}
  \caption{The MLP blocks.}
  \label{fig:MLPblock}
\end{figure*}

Although many Transformer-based architectures such as PatchTST\cite{nietime}, iTranformer\cite{liuitransformer} and DUET\cite{qiu2024duet} have demonstrated that self-attention can be effective for long-term forecasting, recent studies\cite{zeng2023transformers,sun2025simple} argue that sophisticated designs might not be necessary and suggest simple models like feedforward neural networks could suffice for the job. To thoroughly test the TARFVAE framework's effectiveness and avoid confounding effects from elaborately designed temporal processing modules, we implement simple MLP blocks as shown in \textbf{Figure \ref{fig:MLPblock}} for basic input mixing, where the channel-wise mappings enable the capture of inter-channel dependencies. These MLP blocks form the architectural foundation across our encoder, decoder, and prior module.

In the prior module and encoder, following multiple MLP blocks, linear layers output the estimated mean and variance for the Gaussian prior $p(z|x)$ and initial posterior $q(z_0|x,y)$ respectively as follows:
\begin{align}
   \mu_{prior}, \log\sigma_{prior}^2 &= Linear_{prior}(MLPBlocks_{prior}(Emb_{1}(x)))\\
   \mu_{z_0}, \log\sigma_{z_0}^2 &= Linear_{enc}(MLPBlocks_{enc}(Emb_{2}([x,y])))
\end{align}

The decoder aims to reconstruct $y$ using $z$ conditioned on $x$. We chose to augment $z$ via connecting a full attention output where $z$ queries out relevant information from $x$. Subsequently, MLP blocks process this mixed information, and a linear projection is used to achieve the final reconstruction of y. The process is outlined as follows: 
\begin{align}
   h_{mixed} &= z + FullAttention(z,Emb_{4}(x),Emb_{4}(x)) \\
   \hat y &= Linear_{dec}(MLPBlocks_{dec}(h_{mixed}))
\end{align}

\subsection{TARFLOW}
\label{sec:tarflow}
In the flow module, the input $z_0$ and all intermediate latent variables $\{z_k\} (k=1,2,...,K)$ maintain consistent dimensionality $\mathbb{R}^{C\times D}$, where $D$ represents the latent dimension. These variables are naturally partitioned by channels, and we perform the transformation (\ref{eq:tarflow}) conditioned on $x,y$ in each TARFLOW block as follows:
\begin{equation}
\begin{cases}
z_1=x_1\\
z_j= (x_j - t^j((x+Emb_3([x,y]))_{<j}))\odot \exp(s^j((x+Emb_3([x,y]))_{<j}))
\end{cases}
\label{eq:modifiedtarflow}
\end{equation}
Following the original implementation\cite{zhai2024normalizing}, we adopt a dimension-reversing permutation between adjacent blocks. This module finally outputs $z_K \approx z$.

Combining (\ref{eq:logdet}), (\ref{eq:flowloss}) and (\ref{eq:cvaeloss}), the training loss for a single sample generation is derived as
\begin{align}
\begin{split}
L&=-\mathbb{E}_{z \sim q(z|x,y)}\left[\log p(y|z,x)- \log{q(z_0|x,y)}+\sum_{k=1}^{K} \log|\det\frac{\partial z_k}{\partial z_{k-1}}  |+\log{p(z|x)}\right]
\end{split}
 \\
\begin{split}
&= \frac{D}{2}\text{ln}(2\pi)+\frac{1}{2}\sum_{d=1}^{D}{(\text{ln}\sigma_{y,d}^2 - \text{ln}\sigma_{z_{0},d}^2 + \text{ln}\sigma_{prior,d}^2)} - \sum_{k=1}^{K} \sum_{d=1}^D s_k^d \\
&\quad + \frac{1}{2}||\frac{y-\mu_{y}}{\sigma_{y}}||^2 - \frac{1}{2}||\frac{z_0-\mu_{z_{0}}}{\sigma_{z_{0}}}||^2 + \frac{1}{2}||\frac{
z-\mu_{prior}}{\sigma_{prior}}||^2
\end{split}
\label{eq:finalloss}
\end{align}
where $p(y|z,x)$ is chosen to be Guassian for continuous predictions and its mean $\mu_y$ and variance $\sigma_y^2$ are implemented as $\hat y$ and $I$ in our work.

\section{Experiments}
\subsection{Setup}
\label{sec:setup}
\textbf{Datasets}. To comprehensively evaluate the performance of our proposed TARFVAE, we conduct extensive experiments on 8 widely-used real-world datasets: four ETT subsets (ETTh1, ETTh2, ETTm1, ETTm2), Electricity, Exchange, Weather\cite{zhou2021informer,wu2021autoformer},  and Solar-Energy\cite{lai2018modeling}.

\textbf{Baselines}. We extensively choose the recent state-of-the-art models to serve as baselines. For deterministic methods, we include Linear-based or MLP-based methods (SOFTS\cite{hansofts}, TiDE\cite{das2023longterm}, TSMixer\cite{ekambaram2023tsmixer} and DLinear\cite{zeng2023transformers}) and Transformer-based models (DUET\cite{qiu2024duet}, iTransformer\cite{liuitransformer}, PatchTST\cite{nietime}, Crossformer\cite{zhang2023crossformer} and FEDformer\cite{zhou2022fedformer}). For generative approaches, given their distinct evaluation protocols compared to deterministic approaches, we follow mr-Diff\cite{shen2024multi} and compare against both mr-Diff and its benchmarked generative baselines: TimeDiff\cite{shen2023non}, TimeGrad\cite{rasul2021autoregressive}, CSDI\cite{tashiro2021csdi}, SSSD\cite{lopez2023diffusion}, D$^3$VAE\cite{li2022generative}, CPF\cite{rangapuram2023coherent} and PSA-GAN\cite{jeha2022psa}.

\textbf{Implementation details}. When comparing with deterministic models, the long-term forecasting benchmarks follow the common setting\cite{liuitransformer, zhou2021informer,wu2021autoformer,hansofts}, with the lookback window length $L$ set to 96 and the prediction horizon $H$ to $\{96, 192, 336, 720\}$ for all datasets. For comparison with mr-Diff and its benchmarked baselines, we adopt the same configurations: $H$ is 168 for Electricity and ETTh1, 192 for ETTm1, and 672 for Weather, while $L$ is chosen from $\{96, 192, 336, 720, 1440\}$. Mean Squared Error (MSE) and Mean Absolute Error (MAE) are adopted as evaluation metrics. Since we can sample different sizes of results once our model is trained, we calculate MSE and MAE for the median of sampled results. We also compute the Continuous Ranked Probability Score (CRPS)\cite{matheson1976scoring} based on sampled results as a probabilistic forecasting metric. All our experiments are implemented using PyTorch\cite{paszke2019pytorch} on a single Nvidia-H20 GPU with 141 GB memory, except for the inference efficiency comparison experiment which is conducted on an Nvidia-A6000 GPU with 48 GB memory to align with the experimental settings of the compared generative baselines. Our training process is guided by the loss function (\ref{eq:finalloss}) and employs the ADAM optimizer, and the best model is selected based on the MSE of the median of 50 generated samples on the validation set.

\begin{table}[!htbp] 
    \centering
    \caption{Comparison of multivariate time series forecasting results with selected deterministic baselines across various datasets. We reproduce the results for the Exchange dataset and other results of all baselines are taken from DUET\cite{qiu2024duet} and SOFTS\cite{hansofts}.}
    \label{tab:pointestimation}
    \vspace{-3pt}
    \resizebox{\textwidth}{!}{ 
    \begin{tabular}{c|c|c|c|c|c|c|c|c|c|c|c|c|c|c|c|c|c|c|c|c|c}
        \toprule[1.2pt]
        \multicolumn{2}{c}{\multirow{2}{*}{Models}} &
        \multicolumn{2}{c}{TARFVAE} & 
        \multicolumn{2}{c}{DUET} &
        \multicolumn{2}{c}{SOFTS} &
        \multicolumn{2}{c}{iTransformer} &
        \multicolumn{2}{c}{TSMixer} &
        \multicolumn{2}{c}{PatchTST} &
        \multicolumn{2}{c}{Crossformer} &
        \multicolumn{2}{c}{TiDE} &
        \multicolumn{2}{c}{DLinear} &
        \multicolumn{2}{c}{FEDformer} \\
        
        \multicolumn{2}{c}{} &
        \multicolumn{2}{c}{(ours)} & 
        \multicolumn{2}{c}{(2025)} &
        \multicolumn{2}{c}{(2024)} &
        \multicolumn{2}{c}{(2024)} &
        \multicolumn{2}{c}{(2023)} &
        \multicolumn{2}{c}{(2023)} &
        \multicolumn{2}{c}{(2023)} &
        \multicolumn{2}{c}{(2023)} &
        \multicolumn{2}{c}{(2023)} &
        \multicolumn{2}{c}{(2022)} \\
        
        \cmidrule(lr){3-4}
        \cmidrule(lr){5-6} 
        \cmidrule(lr){7-8}
        \cmidrule(lr){9-10} 
        \cmidrule(lr){11-12}
        \cmidrule(lr){13-14} 
        \cmidrule(lr){15-16}
        \cmidrule(lr){17-18} 
        \cmidrule(lr){19-20} 
        \cmidrule(lr){21-22}

        \multicolumn{2}{c}{Metric} 
        & \multicolumn{1}{c}{MSE} & MAE 
        & \multicolumn{1}{c}{MSE} & MAE 
        & \multicolumn{1}{c}{MSE} & MAE 
        & \multicolumn{1}{c}{MSE} & MAE 
        & \multicolumn{1}{c}{MSE} & MAE 
        & \multicolumn{1}{c}{MSE} & MAE
        & \multicolumn{1}{c}{MSE} & MAE 
        & \multicolumn{1}{c}{MSE} & MAE 
        & \multicolumn{1}{c}{MSE} & MAE 
        & \multicolumn{1}{c}{MSE} & MAE \\

        \midrule[1.2pt]
        
        \multirow{5}{*}{\rotatebox{90}{ETTh1}}  
        & 96 
        & \multicolumn{1}{c}{\rb{0.361}} & \rb{0.388} 
        & \multicolumn{1}{c}{0.377} & \bu{0.393} 
        & \multicolumn{1}{c}{0.381} & 0.399 
        & \multicolumn{1}{c}{0.386} & 0.405 
        & \multicolumn{1}{c}{0.401} & 0.412 
        & \multicolumn{1}{c}{0.394} & 0.406 
        & \multicolumn{1}{c}{0.423} & 0.448 
        & \multicolumn{1}{c}{0.479} & 0.464 
        & \multicolumn{1}{c}{0.386} & 0.400 
        & \multicolumn{1}{c}{\bu{0.376}} & 0.419 \\ 
        & 192
        & \multicolumn{1}{c}{\rb{0.410}} & \rb{0.422}
        & \multicolumn{1}{c}{0.429} & \bu{0.425}
        & \multicolumn{1}{c}{0.435} & 0.431
        & \multicolumn{1}{c}{0.441} & 0.436
        & \multicolumn{1}{c}{0.452} & 0.442
        & \multicolumn{1}{c}{0.440} & 0.435
        & \multicolumn{1}{c}{0.471} & 0.474
        & \multicolumn{1}{c}{0.525} & 0.492
        & \multicolumn{1}{c}{0.437} & 0.432
        & \multicolumn{1}{c}{\bu{0.420}} & 0.448 \\ 
        & 336
        & \multicolumn{1}{c}{\rb{0.455}} & \rb{0.445}
        & \multicolumn{1}{c}{0.471} & \bu{0.446}
        & \multicolumn{1}{c}{0.480} & 0.452
        & \multicolumn{1}{c}{0.487} & 0.458
        & \multicolumn{1}{c}{0.492} & 0.463
        & \multicolumn{1}{c}{0.491} & 0.462
        & \multicolumn{1}{c}{0.570} & 0.546
        & \multicolumn{1}{c}{0.565} & 0.515
        & \multicolumn{1}{c}{0.481} & 0.459
        & \multicolumn{1}{c}{\bu{0.459}} & 0.465 \\ 
        & 720
        & \multicolumn{1}{c}{\rb{0.481}} & \rb{0.469}
        & \multicolumn{1}{c}{0.496} & 0.480
        & \multicolumn{1}{c}{0.499} & 0.488
        & \multicolumn{1}{c}{0.503} & 0.491
        & \multicolumn{1}{c}{0.507} & 0.490
        & \multicolumn{1}{c}{\bu{0.487}} & \bu{0.479}
        & \multicolumn{1}{c}{0.653} & 0.621
        & \multicolumn{1}{c}{0.594} & 0.558
        & \multicolumn{1}{c}{0.519} & 0.516
        & \multicolumn{1}{c}{0.506} & 0.507 \\ 
        \cmidrule(lr){2-22}
        & Avg
        & \multicolumn{1}{c}{\rb{0.427}} & \rb{0.431}
        & \multicolumn{1}{c}{0.443} & \bu{0.436}
        & \multicolumn{1}{c}{0.449} & 0.443
        & \multicolumn{1}{c}{0.454} & 0.448
        & \multicolumn{1}{c}{0.463} & 0.452
        & \multicolumn{1}{c}{0.453} & 0.446
        & \multicolumn{1}{c}{0.529} & 0.522
        & \multicolumn{1}{c}{0.541} & 0.507
        & \multicolumn{1}{c}{0.456} & 0.452
        & \multicolumn{1}{c}{\bu{0.440}} & 0.459 \\ 
        \midrule
        
        \multirow{5}{*}{\rotatebox{90}{ETTh2}}  
        & 96
        & \multicolumn{1}{c}{\rb{0.273}} & \rb{0.329}
        & \multicolumn{1}{c}{0.296} & 0.345 
        & \multicolumn{1}{c}{0.297} & 0.347 
        & \multicolumn{1}{c}{0.297} & 0.349 
        & \multicolumn{1}{c}{0.319} & 0.361 
        & \multicolumn{1}{c}{\bu{0.288}} & \bu{0.340} 
        & \multicolumn{1}{c}{0.745} & 0.584 
        & \multicolumn{1}{c}{0.400} & 0.440 
        & \multicolumn{1}{c}{0.333} & 0.387 
        & \multicolumn{1}{c}{0.358} & 0.397 \\ 
        & 192
        & \multicolumn{1}{c}{\rb{0.359}} & \rb{0.382}
        & \multicolumn{1}{c}{\bu{0.368}} & \bu{0.389} 
        & \multicolumn{1}{c}{0.373} & 0.394 
        & \multicolumn{1}{c}{0.380} & 0.400 
        & \multicolumn{1}{c}{0.402} & 0.410 
        & \multicolumn{1}{c}{0.376} & 0.395 
        & \multicolumn{1}{c}{0.877} & 0.656 
        & \multicolumn{1}{c}{0.528} & 0.509 
        & \multicolumn{1}{c}{0.477} & 0.476 
        & \multicolumn{1}{c}{0.429} & 0.439 \\ 
        & 336
        & \multicolumn{1}{c}{\rb{0.391}} & \rb{0.409}
        & \multicolumn{1}{c}{0.411} & \bu{0.422} 
        & \multicolumn{1}{c}{\bu{0.410}} & 0.426 
        & \multicolumn{1}{c}{0.428} & 0.432 
        & \multicolumn{1}{c}{0.444} & 0.446 
        & \multicolumn{1}{c}{0.440} & 0.451 
        & \multicolumn{1}{c}{1.043} & 0.731 
        & \multicolumn{1}{c}{0.643} & 0.571 
        & \multicolumn{1}{c}{0.594} & 0.541 
        & \multicolumn{1}{c}{0.496} & 0.487 \\ 
        & 720
        & \multicolumn{1}{c}{\rb{0.399}} & \rb{0.428}
        & \multicolumn{1}{c}{0.412} & 0.434 
        & \multicolumn{1}{c}{\bu{0.411}} & \bu{0.433} 
        & \multicolumn{1}{c}{0.427} & 0.445 
        & \multicolumn{1}{c}{0.441} & 0.450 
        & \multicolumn{1}{c}{0.436} & 0.453 
        & \multicolumn{1}{c}{1.104} & 0.763 
        & \multicolumn{1}{c}{0.874} & 0.679 
        & \multicolumn{1}{c}{0.831} & 0.657 
        & \multicolumn{1}{c}{0.463} & 0.474 \\ 
        \cmidrule(lr){2-22}
        & Avg
        & \multicolumn{1}{c}{\rb{0.355}} & \rb{0.387}
        & \multicolumn{1}{c}{\bu{0.372}} & \bu{0.398} 
        & \multicolumn{1}{c}{0.373} & 0.400 
        & \multicolumn{1}{c}{0.383} & 0.407 
        & \multicolumn{1}{c}{0.402} & 0.417 
        & \multicolumn{1}{c}{0.385} & 0.410 
        & \multicolumn{1}{c}{0.942} & 0.684 
        & \multicolumn{1}{c}{0.611} & 0.550 
        & \multicolumn{1}{c}{0.559} & 0.515 
        & \multicolumn{1}{c}{0.437} & 0.449 \\ 
        \midrule

        \multirow{5}{*}{\rotatebox{90}{ETTm1}}
        & 96 
        & \multicolumn{1}{c}{\rb{0.311}} & \rb{0.351} 
        & \multicolumn{1}{c}{0.324} & \bu{0.354} 
        & \multicolumn{1}{c}{0.325} & 0.361 
        & \multicolumn{1}{c}{0.334} & 0.368 
        & \multicolumn{1}{c}{\bu{0.323}} & 0.363 
        & \multicolumn{1}{c}{0.329} & 0.365 
        & \multicolumn{1}{c}{0.404} & 0.426 
        & \multicolumn{1}{c}{0.364} & 0.387 
        & \multicolumn{1}{c}{0.345} & 0.372 
        & \multicolumn{1}{c}{0.379} & 0.419 \\ 
        & 192 
        & \multicolumn{1}{c}{\rb{0.361}} & \rb{0.378} 
        & \multicolumn{1}{c}{\bu{0.369}} & \bu{0.379} 
        & \multicolumn{1}{c}{0.375} & 0.389 
        & \multicolumn{1}{c}{0.377} & 0.391 
        & \multicolumn{1}{c}{0.376} & 0.392 
        & \multicolumn{1}{c}{0.380} & 0.394 
        & \multicolumn{1}{c}{0.450} & 0.451 
        & \multicolumn{1}{c}{0.398} & 0.404 
        & \multicolumn{1}{c}{0.380} & 0.389 
        & \multicolumn{1}{c}{0.426} & 0.441 \\ 
        & 336 
        & \multicolumn{1}{c}{\rb{0.391}} & \rb{0.401} 
        & \multicolumn{1}{c}{0.404} & \bu{0.402} 
        & \multicolumn{1}{c}{0.405} & 0.412 
        & \multicolumn{1}{c}{0.426} & 0.420 
        & \multicolumn{1}{c}{0.407} & 0.413 
        & \multicolumn{1}{c}{\bu{0.400}} & 0.410 
        & \multicolumn{1}{c}{0.532} & 0.515 
        & \multicolumn{1}{c}{0.428} & 0.425 
        & \multicolumn{1}{c}{0.413} & 0.413 
        & \multicolumn{1}{c}{0.445} & 0.459 \\ 
        & 720 
        & \multicolumn{1}{c}{\rb{0.456}} & \rb{0.435} 
        & \multicolumn{1}{c}{\bu{0.463}} & \bu{0.437} 
        & \multicolumn{1}{c}{0.466} & 0.447 
        & \multicolumn{1}{c}{0.491} & 0.459 
        & \multicolumn{1}{c}{0.485} & 0.459 
        & \multicolumn{1}{c}{0.475} & 0.453 
        & \multicolumn{1}{c}{0.666} & 0.589 
        & \multicolumn{1}{c}{0.487} & 0.461 
        & \multicolumn{1}{c}{0.474} & 0.453 
        & \multicolumn{1}{c}{0.543} & 0.490 \\
        \cmidrule(lr){2-22}
        & Avg 
        & \multicolumn{1}{c}{\rb{0.380}} & \rb{0.391} 
        & \multicolumn{1}{c}{\bu{0.390}} & \bu{0.393} 
        & \multicolumn{1}{c}{0.393} & 0.402 
        & \multicolumn{1}{c}{0.407} & 0.410 
        & \multicolumn{1}{c}{0.398} & 0.407 
        & \multicolumn{1}{c}{0.396} & 0.406 
        & \multicolumn{1}{c}{0.513} & 0.495 
        & \multicolumn{1}{c}{0.419} & 0.419 
        & \multicolumn{1}{c}{0.403} & 0.407 
        & \multicolumn{1}{c}{0.448} & 0.452 \\ 
        \midrule
        
        \multirow{5}{*}{\rotatebox{90}{ETTm2}}
        & 96 
        & \multicolumn{1}{c}{\rb{0.171}} & \rb{0.249} 
        & \multicolumn{1}{c}{\bu{0.174}} & \bu{0.255} 
        & \multicolumn{1}{c}{0.180} & 0.261 
        & \multicolumn{1}{c}{0.180} & 0.264 
        & \multicolumn{1}{c}{0.182} & 0.266 
        & \multicolumn{1}{c}{0.184} & 0.264 
        & \multicolumn{1}{c}{0.287} & 0.366 
        & \multicolumn{1}{c}{0.207} & 0.305 
        & \multicolumn{1}{c}{0.193} & 0.292 
        & \multicolumn{1}{c}{0.203} & 0.287 \\ 
        & 192 
        & \multicolumn{1}{c}{\rb{0.229}} & \rb{0.291} 
        & \multicolumn{1}{c}{\bu{0.243}} & \bu{0.302} 
        & \multicolumn{1}{c}{0.246} & 0.306 
        & \multicolumn{1}{c}{0.250} & 0.309 
        & \multicolumn{1}{c}{0.249} & 0.309 
        & \multicolumn{1}{c}{0.246} & 0.306 
        & \multicolumn{1}{c}{0.414} & 0.492 
        & \multicolumn{1}{c}{0.290} & 0.364 
        & \multicolumn{1}{c}{0.284} & 0.362 
        & \multicolumn{1}{c}{0.269} & 0.328 \\ 
        & 336 
        & \multicolumn{1}{c}{\rb{0.293}} & \rb{0.334} 
        & \multicolumn{1}{c}{\bu{0.304}} & \bu{0.341} 
        & \multicolumn{1}{c}{0.319} & 0.352 
        & \multicolumn{1}{c}{0.311} & 0.348 
        & \multicolumn{1}{c}{0.309} & 0.347 
        & \multicolumn{1}{c}{0.308} & 0.346 
        & \multicolumn{1}{c}{0.597} & 0.542 
        & \multicolumn{1}{c}{0.377} & 0.422 
        & \multicolumn{1}{c}{0.369} & 0.427 
        & \multicolumn{1}{c}{0.325} & 0.366 \\ 
        & 720 
        & \multicolumn{1}{c}{\rb{0.391}} & \rb{0.391} 
        & \multicolumn{1}{c}{\bu{0.399}} & \bu{0.397} 
        & \multicolumn{1}{c}{0.405} & 0.401 
        & \multicolumn{1}{c}{0.412} & 0.407 
        & \multicolumn{1}{c}{0.416} & 0.408 
        & \multicolumn{1}{c}{0.409} & 0.402 
        & \multicolumn{1}{c}{1.730} & 1.042 
        & \multicolumn{1}{c}{0.558} & 0.524 
        & \multicolumn{1}{c}{0.554} & 0.522 
        & \multicolumn{1}{c}{0.421} & 0.415 \\ 
        \cmidrule(lr){2-22}
        & Avg 
        & \multicolumn{1}{c}{\rb{0.271}} & \rb{0.316} 
        & \multicolumn{1}{c}{\bu{0.280}} & \bu{0.324} 
        & \multicolumn{1}{c}{0.288} & 0.330 
        & \multicolumn{1}{c}{0.288} & 0.332 
        & \multicolumn{1}{c}{0.289} & 0.333 
        & \multicolumn{1}{c}{0.287} & 0.330 
        & \multicolumn{1}{c}{0.757} & 0.611 
        & \multicolumn{1}{c}{0.358} & 0.404 
        & \multicolumn{1}{c}{0.350} & 0.401 
        & \multicolumn{1}{c}{0.305} & 0.349 \\ 
        \midrule
        
        \multirow{5}{*}{\rotatebox{90}{Exchange}}
        & 96 
        & \multicolumn{1}{c}{\rb{0.086}} & \rb{0.204} 
        & \multicolumn{1}{c}{\rb{0.086}} & \bu{0.205} 
        & \multicolumn{1}{c}{0.090} & 0.211 
        & \multicolumn{1}{c}{\rb{0.086}} & 0.206 
        & \multicolumn{1}{c}{0.166} & 0.316 
        & \multicolumn{1}{c}{0.088} & \bu{0.205} 
        & \multicolumn{1}{c}{0.256} & 0.367 
        & \multicolumn{1}{c}{0.094} & 0.218 
        & \multicolumn{1}{c}{0.088} & 0.218 
        & \multicolumn{1}{c}{0.148} & 0.278 \\ 
        & 192 
        & \multicolumn{1}{c}{\rb{0.172}} & \rb{0.294} 
        & \multicolumn{1}{c}{0.182} & 0.305 
        & \multicolumn{1}{c}{0.182} & 0.304 
        & \multicolumn{1}{c}{0.177} & \bu{0.299} 
        & \multicolumn{1}{c}{0.279} & 0.402 
        & \multicolumn{1}{c}{\bu{0.176}} & \bu{0.299} 
        & \multicolumn{1}{c}{0.470} & 0.509 
        & \multicolumn{1}{c}{0.184} & 0.307 
        & \multicolumn{1}{c}{\bu{0.176}} & 0.315 
        & \multicolumn{1}{c}{0.271} & 0.315 \\ 
        & 336 
        & \multicolumn{1}{c}{0.340} & 0.423 
        & \multicolumn{1}{c}{\bu{0.310}} & \bu{0.403} 
        & \multicolumn{1}{c}{0.363} & 0.438 
        & \multicolumn{1}{c}{0.331} & 0.417 
        & \multicolumn{1}{c}{0.477} & 0.548 
        & \multicolumn{1}{c}{\rb{0.301}} & \rb{0.397} 
        & \multicolumn{1}{c}{1.268} & 0.883 
        & \multicolumn{1}{c}{0.349} & 0.431 
        & \multicolumn{1}{c}{0.313} & 0.427 
        & \multicolumn{1}{c}{0.460} & 0.427 \\ 
        & 720 
        & \multicolumn{1}{c}{0.797} & 0.678 
        & \multicolumn{1}{c}{\bu{0.693}} & \rb{0.624}
        & \multicolumn{1}{c}{0.997} & 0.743 
        & \multicolumn{1}{c}{0.847} & 0.691 
        & \multicolumn{1}{c}{\rb{0.654}} & \bu{0.662} 
        & \multicolumn{1}{c}{0.901} & 0.714 
        & \multicolumn{1}{c}{1.767} & 1.068 
        & \multicolumn{1}{c}{0.852} & 0.698 
        & \multicolumn{1}{c}{0.839} & 0.695 
        & \multicolumn{1}{c}{1.195} & 0.695 \\ 
        \cmidrule(lr){2-22}
        & Avg 
        & \multicolumn{1}{c}{\bu{0.349}} & \bu{0.400} 
        & \multicolumn{1}{c}{\rb{0.318}} & \rb{0.384} 
        & \multicolumn{1}{c}{0.408} & 0.424 
        & \multicolumn{1}{c}{0.360} & 0.403 
        & \multicolumn{1}{c}{0.394} & 0.482 
        & \multicolumn{1}{c}{0.367} & 0.404 
        & \multicolumn{1}{c}{0.940} & 0.707 
        & \multicolumn{1}{c}{0.370} & 0.414 
        & \multicolumn{1}{c}{0.354} & 0.414 
        & \multicolumn{1}{c}{0.519} & 0.429 \\ 
        \midrule
        
        \multirow{5}{*}{\rotatebox{90}{Electricity}}
        & 96 
        & \multicolumn{1}{c}{\rb{0.139}} & 0.238 
        & \multicolumn{1}{c}{0.145} & \rb{0.233} 
        & \multicolumn{1}{c}{\bu{0.143}} & \rb{0.233} 
        & \multicolumn{1}{c}{0.148} & 0.240 
        & \multicolumn{1}{c}{0.157} & 0.260 
        & \multicolumn{1}{c}{0.164} & 0.251 
        & \multicolumn{1}{c}{0.219} & 0.314 
        & \multicolumn{1}{c}{0.237} & 0.329 
        & \multicolumn{1}{c}{0.197} & 0.282 
        & \multicolumn{1}{c}{0.193} & 0.308 \\ 
        & 192 
        & \multicolumn{1}{c}{\bu{0.159}} & 0.255 
        & \multicolumn{1}{c}{0.163} & \rb{0.248} 
        & \multicolumn{1}{c}{\rb{0.158}} & \rb{0.248} 
        & \multicolumn{1}{c}{0.162} & 0.253 
        & \multicolumn{1}{c}{0.173} & 0.274 
        & \multicolumn{1}{c}{0.173} & 0.262 
        & \multicolumn{1}{c}{0.231} & 0.322 
        & \multicolumn{1}{c}{0.236} & 0.330 
        & \multicolumn{1}{c}{0.196} & 0.285 
        & \multicolumn{1}{c}{0.201} & 0.315 \\ 
        & 336 
        & \multicolumn{1}{c}{\rb{0.173}} & 0.270 
        & \multicolumn{1}{c}{\bu{0.175}} & \rb{0.262} 
        & \multicolumn{1}{c}{0.178} & \bu{0.269} 
        & \multicolumn{1}{c}{0.178} & \bu{0.269} 
        & \multicolumn{1}{c}{0.192} & 0.295 
        & \multicolumn{1}{c}{0.190} & 0.279 
        & \multicolumn{1}{c}{0.246} & 0.337 
        & \multicolumn{1}{c}{0.249} & 0.344 
        & \multicolumn{1}{c}{0.209} & 0.301 
        & \multicolumn{1}{c}{0.214} & 0.329 \\ 
        & 720 
        & \multicolumn{1}{c}{\rb{0.202}} & \bu{0.294} 
        & \multicolumn{1}{c}{\bu{0.204}} & \rb{0.291} 
        & \multicolumn{1}{c}{0.218} & 0.305 
        & \multicolumn{1}{c}{0.225} & 0.317 
        & \multicolumn{1}{c}{0.223} & 0.318 
        & \multicolumn{1}{c}{0.230} & 0.313 
        & \multicolumn{1}{c}{0.280} & 0.363 
        & \multicolumn{1}{c}{0.284} & 0.373 
        & \multicolumn{1}{c}{0.245} & 0.333 
        & \multicolumn{1}{c}{0.246} & 0.355 \\ 
        \cmidrule(lr){2-22}
        & Avg 
        & \multicolumn{1}{c}{\rb{0.168}} & \bu{0.264} 
        & \multicolumn{1}{c}{\bu{0.172}} & \rb{0.259} 
        & \multicolumn{1}{c}{0.174} & \bu{0.264} 
        & \multicolumn{1}{c}{0.178} & 0.270 
        & \multicolumn{1}{c}{0.186} & 0.287 
        & \multicolumn{1}{c}{0.189} & 0.276 
        & \multicolumn{1}{c}{0.244} & 0.334 
        & \multicolumn{1}{c}{0.252} & 0.344 
        & \multicolumn{1}{c}{0.212} & 0.300 
        & \multicolumn{1}{c}{0.214} & 0.327 \\ 
        \midrule
        
        \multirow{5}{*}{\rotatebox{90}{Solar}} 
        & 96 
        & \multicolumn{1}{c}{\rb{0.195}} & \bu{0.220} 
        & \multicolumn{1}{c}{\bu{0.200}} & \rb{0.207} 
        & \multicolumn{1}{c}{\bu{0.200}} & 0.230 
        & \multicolumn{1}{c}{0.203} & 0.237 
        & \multicolumn{1}{c}{0.221} & 0.275 
        & \multicolumn{1}{c}{0.205} & 0.246 
        & \multicolumn{1}{c}{0.310} & 0.331 
        & \multicolumn{1}{c}{0.312} & 0.399 
        & \multicolumn{1}{c}{0.290} & 0.378 
        & \multicolumn{1}{c}{0.242} & 0.342 \\ 
        & 192 
        & \multicolumn{1}{c}{\rb{0.225}} & \bu{0.242} 
        & \multicolumn{1}{c}{\bu{0.228}} & \rb{0.233} 
        & \multicolumn{1}{c}{0.229} & 0.253 
        & \multicolumn{1}{c}{0.233} & 0.261 
        & \multicolumn{1}{c}{0.268} & 0.306 
        & \multicolumn{1}{c}{0.237} & 0.267 
        & \multicolumn{1}{c}{0.734} & 0.725 
        & \multicolumn{1}{c}{0.339} & 0.416 
        & \multicolumn{1}{c}{0.320} & 0.398 
        & \multicolumn{1}{c}{0.285} & 0.380 \\ 
        & 336 
        & \multicolumn{1}{c}{0.250} & \bu{0.262} 
        & \multicolumn{1}{c}{0.262} & \rb{0.244} 
        & \multicolumn{1}{c}{\rb{0.243}} & 0.269 
        & \multicolumn{1}{c}{\bu{0.248}} & 0.273 
        & \multicolumn{1}{c}{0.272} & 0.294 
        & \multicolumn{1}{c}{0.250} & 0.276 
        & \multicolumn{1}{c}{0.750} & 0.735 
        & \multicolumn{1}{c}{0.368} & 0.430 
        & \multicolumn{1}{c}{0.353} & 0.415 
        & \multicolumn{1}{c}{0.282} & 0.376 \\ 
        & 720 
        & \multicolumn{1}{c}{0.254} & \bu{0.269} 
        & \multicolumn{1}{c}{0.258} & \rb{0.249} 
        & \multicolumn{1}{c}{\rb{0.245}} & 0.272 
        & \multicolumn{1}{c}{\bu{0.249}} & 0.275 
        & \multicolumn{1}{c}{0.281} & 0.313 
        & \multicolumn{1}{c}{0.252} & 0.275 
        & \multicolumn{1}{c}{0.769} & 0.765 
        & \multicolumn{1}{c}{0.370} & 0.425 
        & \multicolumn{1}{c}{0.356} & 0.413 
        & \multicolumn{1}{c}{0.357} & 0.427 \\ 
        \cmidrule(lr){2-22}
        & Avg 
        & \multicolumn{1}{c}{\bu{0.231}} & \bu{0.248} 
        & \multicolumn{1}{c}{0.237} & \rb{0.233} 
        & \multicolumn{1}{c}{\rb{0.229}} & 0.256 
        & \multicolumn{1}{c}{0.233} & 0.262 
        & \multicolumn{1}{c}{0.261} & 0.297 
        & \multicolumn{1}{c}{0.236} & 0.266 
        & \multicolumn{1}{c}{0.641} & 0.639 
        & \multicolumn{1}{c}{0.347} & 0.418 
        & \multicolumn{1}{c}{0.330} & 0.401 
        & \multicolumn{1}{c}{0.292} & 0.381 \\ 
        \midrule
        
        \multirow{5}{*}{\rotatebox{90}{Weather}}
        & 96 
        & \multicolumn{1}{c}{\rb{0.151}} & \rb{0.197} 
        & \multicolumn{1}{c}{0.163} & \bu{0.202} 
        & \multicolumn{1}{c}{0.166} & 0.208 
        & \multicolumn{1}{c}{0.174} & 0.214 
        & \multicolumn{1}{c}{0.166} & 0.210 
        & \multicolumn{1}{c}{0.176} & 0.217 
        & \multicolumn{1}{c}{\bu{0.158}} & 0.230 
        & \multicolumn{1}{c}{0.202} & 0.261 
        & \multicolumn{1}{c}{0.196} & 0.255 
        & \multicolumn{1}{c}{0.217} & 0.296 \\ 
        & 192 
        & \multicolumn{1}{c}{\rb{0.202}} & \rb{0.242} 
        & \multicolumn{1}{c}{0.218} & \bu{0.252} 
        & \multicolumn{1}{c}{0.217} & 0.253 
        & \multicolumn{1}{c}{0.221} & 0.254 
        & \multicolumn{1}{c}{0.215} & 0.256 
        & \multicolumn{1}{c}{0.221} & 0.256 
        & \multicolumn{1}{c}{\bu{0.206}} & 0.277 
        & \multicolumn{1}{c}{0.242} & 0.298 
        & \multicolumn{1}{c}{0.237} & 0.296 
        & \multicolumn{1}{c}{0.276} & 0.336 \\ 
        & 336 
        & \multicolumn{1}{c}{\rb{0.262}} & \rb{0.289} 
        & \multicolumn{1}{c}{0.274} & \bu{0.294} 
        & \multicolumn{1}{c}{0.282} & 0.300 
        & \multicolumn{1}{c}{0.278} & 0.296 
        & \multicolumn{1}{c}{0.287} & 0.300 
        & \multicolumn{1}{c}{0.275} & 0.296 
        & \multicolumn{1}{c}{\bu{0.272}} & 0.335 
        & \multicolumn{1}{c}{0.287} & 0.335 
        & \multicolumn{1}{c}{0.283} & 0.335 
        & \multicolumn{1}{c}{0.339} & 0.380 \\ 
        & 720 
        & \multicolumn{1}{c}{\rb{0.342}} & \rb{0.341} 
        & \multicolumn{1}{c}{\bu{0.349}} & \bu{0.343} 
        & \multicolumn{1}{c}{0.356} & 0.351 
        & \multicolumn{1}{c}{0.358} & 0.347 
        & \multicolumn{1}{c}{0.355} & 0.348 
        & \multicolumn{1}{c}{0.352} & 0.346 
        & \multicolumn{1}{c}{0.398} & 0.418 
        & \multicolumn{1}{c}{0.351} & 0.386 
        & \multicolumn{1}{c}{0.345} & 0.381 
        & \multicolumn{1}{c}{0.403} & 0.428 \\ 
        \cmidrule(lr){2-22}
        & Avg 
        & \multicolumn{1}{c}{\rb{0.239}} & \rb{0.267} 
        & \multicolumn{1}{c}{\bu{0.251}} & \bu{0.273} 
        & \multicolumn{1}{c}{0.255} & 0.278 
        & \multicolumn{1}{c}{0.258} & 0.278 
        & \multicolumn{1}{c}{0.256} & 0.279 
        & \multicolumn{1}{c}{0.256} & 0.279 
        & \multicolumn{1}{c}{0.259} & 0.315 
        & \multicolumn{1}{c}{0.271} & 0.320 
        & \multicolumn{1}{c}{0.265} & 0.317 
        & \multicolumn{1}{c}{0.309} & 0.360 \\ 

        \midrule

        \multicolumn{2}{l|}{$1^{st}$ Count}
        & \multicolumn{1}{c}{33} & 27 
        & \multicolumn{1}{c}{2} & 12 
        & \multicolumn{1}{c}{4} & 2 
        & \multicolumn{1}{c}{1} & 0
        & \multicolumn{1}{c}{1} & 0
        & \multicolumn{1}{c}{1} & 1 
        & \multicolumn{1}{c}{0} & 0 
        & \multicolumn{1}{c}{0} & 0
        & \multicolumn{1}{c}{0} & 0
        & \multicolumn{1}{c}{0} & 0 \\         

        \bottomrule[1.2pt]
    \end{tabular}
    }

\end{table}

\begin{table}[!ht]
    \centering
    \caption{The inference results of TARFVAE with varing sample sizes: $\{20,50,100,200\}$. The CRPS is calculated using the method in Appendix \ref{apd:crpsmethod}.}
    \vspace{-5pt}
    \resizebox{\textwidth}{!}{
    \label{tab:crps}
    \begin{tabular}{c|c|ccc|ccc|ccc|ccc|ccc|ccc|ccc|ccc}

        \toprule[1.2pt]
        
        \multicolumn{2}{c}{Datasets} &
        \multicolumn{3}{c}{ETTh1} & 
        \multicolumn{3}{c}{ETTh2} &
        \multicolumn{3}{c}{ETTm1} & 
        \multicolumn{3}{c}{ETTm2} &
        \multicolumn{3}{c}{Exchange} & 
        \multicolumn{3}{c}{Electricity} &
        \multicolumn{3}{c}{Solar} & 
        \multicolumn{3}{c}{Weather} \\

        \cmidrule(lr){3-5}
        \cmidrule(lr){6-8} 
        \cmidrule(lr){9-11}
        \cmidrule(lr){12-14} 
        \cmidrule(lr){15-17}
        \cmidrule(lr){18-20} 
        \cmidrule(lr){21-23}
        \cmidrule(lr){24-26}

        \multicolumn{2}{c}{Metric} &
        \multicolumn{1}{c}{MSE} & \multicolumn{1}{c}{MAE} & CRPS &
        \multicolumn{1}{c}{MSE} & \multicolumn{1}{c}{MAE} & CRPS &
        \multicolumn{1}{c}{MSE} & \multicolumn{1}{c}{MAE} & CRPS &
        \multicolumn{1}{c}{MSE} & \multicolumn{1}{c}{MAE} & CRPS &
        \multicolumn{1}{c}{MSE} & \multicolumn{1}{c}{MAE} & CRPS &
        \multicolumn{1}{c}{MSE} & \multicolumn{1}{c}{MAE} & CRPS &
        \multicolumn{1}{c}{MSE} & \multicolumn{1}{c}{MAE} & CRPS &
        \multicolumn{1}{c}{MSE} & \multicolumn{1}{c}{MAE} & CRPS \\

        \midrule

        \multirow{4}{*}{96} & 20 & 
        0.376 & 0.399 & 0.323 & 
        0.275 & 0.331 & 0.302 & 
        0.321 & 0.359 & 0.296 & 
        0.177 & 0.257 & 0.215 & 
        0.087 & 0.206 & 0.176 & 
        0.143 & 0.243 & 0.203 & 
        0.201 & 0.224 & 0.183 & 
        0.154 & 0.200 & 0.173 \\ 
        
        & 50 & 
        0.366 & 0.391 & 0.315 & 
        0.274 & 0.330 & 0.298 & 
        0.314 & 0.354 & 0.289 & 
        0.173 & 0.252 & 0.210 & 
        \rb{0.086} & \rb{0.204} & 0.173 & 
        0.141 & 0.240 & 0.198 & 
        0.197 & 0.221 & 0.179 & 
        0.152 & 0.198 & 0.170 \\

        & 100 & 
        0.363 & 0.389 & 0.313 & 
        \rb{0.273} & \rb{0.329} & 0.297 & 
        0.312 & 0.352 & 0.287 & 
        \rb{0.171} & 0.250 & 0.209 & 
        \rb{0.086} & \rb{0.204} & 0.172 & 
        0.140 & 0.239 & 0.197 & 
        0.196 & \rb{0.220} & \rb{0.177} & 
        \rb{0.151} & 0.198 & \rb{0.169} \\

        & 200 & 
        \rb{0.361} & \rb{0.388} & \rb{0.311} & 
        \rb{0.273} & \rb{0.329} & \rb{0.296} & 
        \rb{0.311} & \rb{0.351} & \rb{0.286} & 
        \rb{0.171} & \rb{0.249} & \rb{0.208} & 
        \rb{0.086} & \rb{0.204} & \rb{0.171} & 
        \rb{0.139} & \rb{0.238} & \rb{0.196} & 
        \rb{0.195} & \rb{0.220} & \rb{0.177} & 
        \rb{0.151} & \rb{0.197} & \rb{0.169} \\

        \midrule

        \multirow{4}{*}{192} & 20 & 
        0.423 & 0.430 & 0.355 & 
        0.361 & 0.384 & 0.355 & 
        0.373 & 0.387 & 0.321 & 
        0.233 & 0.296 & 0.254 & 
        0.176 & 0.298 & 0.254 & 
        0.163 & 0.260 & 0.219 & 
        0.230 & 0.245 & 0.207 & 
        0.207 & 0.246 & 0.210 \\

        & 50 & 
        0.415 & 0.425 & 0.347 & 
        \rb{0.359} & \rb{0.382} & 0.351 & 
        0.365 & 0.381 & 0.313 & 
        0.230 & 0.293 & 0.249 & 
        0.173 & 0.295 & 0.248 & 
        0.160 & 0.257 & 0.215 & 
        0.227 & 0.243 & 0.202 & 
        0.204 & 0.243 & 0.206 \\

        & 100 & 
        0.412 & 0.423 & 0.344 & 
        \rb{0.359} & \rb{0.382} & 0.350 & 
        0.363 & 0.379 & 0.311 & 
        \rb{0.229} & 0.292 & \rb{0.247} & 
        \rb{0.172} & \rb{0.294} & 0.246 & 
        \rb{0.159} & 0.256 & 0.213 & 
        0.226 & 0.243 & 0.201 & 
        0.203 & \rb{0.242} & 0.204 \\

        & 200 & 
        \rb{0.410} & \rb{0.422} & \rb{0.342} & 
        \rb{0.359} & \rb{0.382} & \rb{0.349} & 
        \rb{0.361} & \rb{0.378} & \rb{0.309} & 
        \rb{0.229} & \rb{0.291} & \rb{0.247} & 
        \rb{0.172} & \rb{0.294} & \rb{0.245} & 
        \rb{0.159} & \rb{0.255} & \rb{0.212} & 
        \rb{0.225} & \rb{0.242} & \rb{0.200} & 
        \rb{0.202} & \rb{0.242} & \rb{0.203} \\

        \midrule

        \multirow{4}{*}{336} & 20 & 
        0.465 & 0.452 & 0.385 & 
        0.393 & 0.412 & 0.380 & 
        0.404 & 0.411 & 0.340 & 
        0.301 & 0.339 & 0.293 & 
        \rb{0.338} & \rb{0.422} & 0.381 & 
        0.179 & 0.276 & 0.230 & 
        0.254 & 0.266 & 0.225 & 
        0.269 & 0.293 & 0.252 \\

        & 50 & 
        0.458 & 0.447 & 0.378 & 
        \rb{0.391} & 0.410 & 0.375 & 
        0.395 & 0.404 & 0.332 & 
        0.295 & 0.335 & 0.288 & 
        0.339 & \rb{0.422} & 0.376 & 
        0.175 & 0.272 & 0.224 & 
        0.252 & 0.263 & 0.220 & 
        0.264 & 0.290 & 0.246 \\

        & 100 & 
        0.456 & 0.446 & 0.375 & 
        \rb{0.391} & 0.410 & 0.373 & 
        0.392 & 0.402 & 0.329 & 
        0.294 & \rb{0.334} & 0.286 & 
        0.340 & 0.423 & 0.374 & 
        0.174 & 0.271 & 0.223 & 
        0.251 & \rb{0.262} & 0.219 & 
        0.263 & \rb{0.289} & \rb{0.244} \\

        & 200 & 
        \rb{0.455} & \rb{0.445} & \rb{0.374} & 
        \rb{0.391} & \rb{0.409} & \rb{0.372} & 
        \rb{0.391} & \rb{0.401} & \rb{0.328} & 
        \rb{0.293} & \rb{0.334} & \rb{0.285} & 
        0.340 & 0.423 & \rb{0.373} & 
        \rb{0.173} & \rb{0.270} & \rb{0.222} & 
        \rb{0.250} & \rb{0.262} & \rb{0.218} & 
        \rb{0.262} & \rb{0.289} & \rb{0.244} \\

        \midrule

        \multirow{4}{*}{720} & 20 & 
        0.502 & 0.483 & 0.414 & 
        0.403 & 0.431 & 0.394 & 
        0.470 & 0.445 & 0.373 & 
        0.400 & 0.397 & 0.345 & 
        \rb{0.792} & \rb{0.676} & 0.647 & 
        0.208 & 0.299 & 0.253 & 
        0.258 & 0.272 & 0.234 & 
        0.349 & 0.345 & 0.300 \\

        & 50 & 
        0.488 & 0.473 & 0.406 & 
        0.400 & 0.429 & 0.389 & 
        0.461 & 0.439 & 0.365 & 
        0.394 & 0.393 & 0.339 & 
        0.796 & 0.678 & 0.640 & 
        0.204 & 0.295 & 0.247 & 
        0.256 & 0.270 & 0.230 & 
        0.344 & 0.342 & 0.295 \\

        & 100 & 
        0.483 & 0.471 & 0.403 & 
        \rb{0.399} & \rb{0.428} & 0.388 & 
        0.458 & 0.437 & 0.362 & 
        0.393 & \rb{0.391} & 0.336 & 
        0.796 & 0.678 & 0.637 & 
        0.203 & \rb{0.294} & 0.246 & 
        0.255 & \rb{0.269} & 0.228 & 
        0.343 & \rb{0.341} & 0.293 \\

        & 200 & 
        \rb{0.481} & \rb{0.469} & \rb{0.401} & 
        \rb{0.399} & \rb{0.428} & \rb{0.387} & 
        \rb{0.456} & \rb{0.435} & \rb{0.360} & 
        \rb{0.391} & \rb{0.391} & \rb{0.335} & 
        0.797 & 0.678 & \rb{0.636} & 
        \rb{0.202} & \rb{0.294} & \rb{0.245} & 
        \rb{0.254} & \rb{0.269} & \rb{0.227} & 
        \rb{0.342} & \rb{0.341} & \rb{0.292} \\

        \midrule

        \multirow{4}{*}{Avg} & 20 & 
        0.442 & 0.441 & 0.369 & 
        0.358 & 0.390 & 0.358 & 
        0.392 & 0.400 & 0.332 & 
        0.278 & 0.322 & 0.277 & 
        \rb{0.348} & \rb{0.400} & 0.365 & 
        0.173 & 0.269 & 0.226 & 
        0.236 & 0.252 & 0.212 & 
        0.245 & 0.271 & 0.234 \\

        & 50 & 
        0.432 & 0.434 & 0.361 & 
        0.356 & 0.388 & 0.353 & 
        0.384 & 0.394 & 0.325 & 
        0.273 & 0.318 & 0.271 & 
        0.349 & \rb{0.400} & 0.359 & 
        0.170 & 0.266 & 0.221 & 
        0.233 & 0.249 & 0.208 & 
        0.241 & 0.268 & 0.229 \\

        & 100 & 
        0.428 & 0.432 & 0.359 & 
        0.356 & \rb{0.387} & 0.352 & 
        0.381 & 0.392 & 0.322 & 
        0.272 & 0.317 & 0.270 & 
        0.349 & \rb{0.400} & 0.357 & 
        0.169 & 0.265 & \rb{0.219} & 
        0.232 & 0.249 & \rb{0.206} & 
        0.240 & \rb{0.267} & 0.228 \\

        & 200 & 
        \rb{0.427} & \rb{0.431} & \rb{0.357} & 
        \rb{0.355} & \rb{0.387} & \rb{0.351} & 
        \rb{0.380} & \rb{0.391} & \rb{0.321} & 
        \rb{0.271} & \rb{0.316} & \rb{0.269} & 
        0.349 & \rb{0.400} & \rb{0.356} & 
        \rb{0.168} & \rb{0.264} & \rb{0.219} & 
        \rb{0.231} & \rb{0.248} & \rb{0.206} & 
        \rb{0.239} & \rb{0.267} & \rb{0.227} \\

        \bottomrule[1.2pt]

    \end{tabular}
    }
    
\end{table}
\vspace{-5pt}

\subsection{Main results}
\textbf{Table \ref{tab:pointestimation}} presents the multivariate time series forecasting performance of TARFVAE in contrast with selected deterministic baselines, where the results of sampling 200 are shown and the full results of different sample sizes can be seen in \textbf{Table \ref{tab:crps}}. The results demonstrate that TARFVAE achieves the highest number of best outcomes across various forecast horizons in all 8 datasets. Notably, TARFVAE consistently achieves superior performance under varing forecast horizons in 5 out of 8 datasets with significant improvements. For instance, it reduces average MSE by $4.3\%$ on ETTh2 and $4.8\%$ on Weather, respectively. These results not only demonstrate TARFVAE's precision advantages compared to state-of-the-art deterministic methods but also reveal its ability to produce robust performance in considerable long-term forecasting (e.g., 720-horizon predictions).

\textbf{Table \ref{tab:peekingpointestimation}} shows the performance of TARFVAE compared to other generative methods. TARFVAE achieves the best overall performance, securing top-1 positions in 5 out of 10 metrics and ranking second in 2 metrics. To compare inference efficiency, we rerun TARFVAE under the same configurations as the baselines, since some baselines such as mr-Diff and TimeDiff are not open-sourced. The results shown in \textbf{Table \ref{tab:inferencetime}} demonstrate TARFVAE's significant speed advantages: for instance, it reduces inference time by $85.6\%$ compared to mr-Diff at horizon $H=720$ in the ETTh1 dataset, and even more so when compared to other methods. Furthermore, TARFVAE maintains computational stability when handling extended forecasting horizons – only the final linear layer's dimensionality increases, while other generative methods require progressive time increments. Additionally, its parallelized sampling computation ensures minimal latency variation (from 1 to 200 samples) under moderate sampling scales. These findings collectively establish TARFVAE as both an accurate and efficient approach, enabling high-quality probabilistic forecasts through one-step generation while outperforming state-of-the-art generative methods in computational scalability.

\begin{table}[htbp] 
    \centering
    \caption{Comparison of multivariate time series forecasting results with selected generative baselines across various datasets. The results of all baselines are taken from mr-Diff\cite{shen2024multi}.}
    \resizebox{0.95\textwidth}{!}{ 
    \label{tab:peekingpointestimation}
    \vspace{-3pt}
    \begin{tabular}{c|l|l|l|l|l|l|l|l|l|l|c}

        \toprule[1.2pt]
        \multicolumn{1}{c}{Datasets} &
        \multicolumn{2}{c}{ETTh1} & 
        \multicolumn{2}{c}{ETTm1} &
        \multicolumn{2}{c}{Exchange} &
        \multicolumn{2}{c}{Electricity} &
        \multicolumn{2}{c}{Weather} & \\

        \cmidrule(lr){2-3}
        \cmidrule(lr){4-5} 
        \cmidrule(lr){6-7}
        \cmidrule(lr){8-9} 
        \cmidrule(lr){10-11}
        \cmidrule(lr){12-12}

        \multicolumn{1}{c}{Metric} 
        & \multicolumn{1}{c}{MSE} & \multicolumn{1}{c}{MAE} 
        & \multicolumn{1}{c}{MSE} & \multicolumn{1}{c}{MAE} 
        & \multicolumn{1}{c}{MSE} & \multicolumn{1}{c}{MAE} 
        & \multicolumn{1}{c}{MSE} & \multicolumn{1}{c}{MAE} 
        & \multicolumn{1}{c}{MSE} & \multicolumn{1}{c}{MAE} 
        & Avg Rank \\

        \midrule

        TARFVAE(ours) 
        & \multicolumn{1}{l}{ \rb{0.405} $_{(1)}$ } & \rb{0.415} $_{(1)}$ 
        & \multicolumn{1}{l}{ \rb{0.321} $_{(1)}$ } & \rb{0.363} $_{(1)}$ 
        & \multicolumn{1}{l}{ \rb{0.016} $_{(1)}$ } & 0.084 $_{(5)}$ 
        & \multicolumn{1}{l}{ \bu{0.150} $_{(2)}$ } & 0.249 $_{(3)}$ 
        & \multicolumn{1}{l}{ \bu{0.309} $_{(2)}$ } & 0.334 $_{(4)}$ & 2.1 \\ 

        mr-Diff 
        & \multicolumn{1}{l}{ 0.411 $_{(4)}$ } & 0.422 $_{(3)}$ 
        & \multicolumn{1}{l}{ 0.340 $_{(3)}$ } & 0.373 $_{(3)}$ 
        & \multicolumn{1}{l}{ \rb{0.016} $_{(1)}$ } & 0.082 $_{(3)}$ 
        & \multicolumn{1}{l}{ 0.155 $_{(4)}$ } & 0.252 $_{(4)}$ 
        & \multicolumn{1}{l}{ \rb{0.296} $_{(1)}$ } & \bu{0.324} $_{(2)}$ & 2.8 \\ 

        TimeDiff 
        & \multicolumn{1}{l}{ \bu{0.407} $_{(2)}$ } & 0.430 $_{(4)}$ 
        & \multicolumn{1}{l}{ \bu{0.336} $_{(2)}$ } & \bu{0.372} $_{(2)}$ 
        & \multicolumn{1}{l}{ 0.018 $_{(7)}$ } & 0.091 $_{(8)}$ 
        & \multicolumn{1}{l}{ 0.193 $_{(6)}$ } & 0.305 $_{(6)}$ 
        & \multicolumn{1}{l}{ 0.311 $_{(3)}$ } & \rb{0.312} $_{(1)}$ & 4.1 \\ 

        TimeGrad 
        & \multicolumn{1}{l}{ 0.993 $_{(16)}$ } & 0.719 $_{(16)}$ 
        & \multicolumn{1}{l}{ 0.874 $_{(16)}$ } & 0.605 $_{(16)}$ 
        & \multicolumn{1}{l}{ 0.079 $_{(15)}$ } & 0.193 $_{(14)}$ 
        & \multicolumn{1}{l}{ 0.736 $_{(15)}$ } & 0.630 $_{(15)}$ 
        & \multicolumn{1}{l}{ 0.392 $_{(11)}$ } & 0.381 $_{(11)}$ & 14.5 \\ 

        \midrule
        
        CSDI 
        & \multicolumn{1}{l}{ 0.497 $_{(8)}$ } & 0.438 $_{(6)}$ 
        & \multicolumn{1}{l}{ 0.529 $_{(14)}$ } & 0.442 $_{(13)}$ 
        & \multicolumn{1}{l}{ 0.077 $_{(14)}$ } & 0.194 $_{(15)}$ 
        & \multicolumn{1}{c}{-} & \multicolumn{1}{c}{-}
        & \multicolumn{1}{l}{ 0.356 $_{(9)}$ } & 0.374 $_{(9)}$ & 11 \\ 

        SSSD 
        & \multicolumn{1}{l}{ 0.726 $_{(14)}$ } & 0.561 $_{(14)}$ 
        & \multicolumn{1}{l}{ 0.464 $_{(12)}$ } & 0.406 $_{(10)}$ 
        & \multicolumn{1}{l}{ 0.061 $_{(13)}$ } & 0.127 $_{(12)}$ 
        & \multicolumn{1}{l}{ 0.255 $_{(9)}$ } & 0.363 $_{(9)}$ 
        & \multicolumn{1}{l}{ 0.349 $_{(8)}$ } & 0.350 $_{(8)}$ & 10.9 \\ 

        D3VAE 
        & \multicolumn{1}{l}{ 0.504 $_{(10)}$ } & 0.502 $_{(11)}$ 
        & \multicolumn{1}{l}{ 0.362 $_{(9)}$ } & 0.391 $_{(9)}$ 
        & \multicolumn{1}{l}{ 0.200 $_{(16)}$ } & 0.301 $_{(16)}$ 
        & \multicolumn{1}{l}{ 0.286 $_{(11)}$ } & 0.372 $_{(11)}$ 
        & \multicolumn{1}{l}{ 0.375 $_{(10)}$ } & 0.380 $_{(10)}$ & 11.3 \\ 

        CPF 
        & \multicolumn{1}{l}{ 0.730 $_{(15)}$ } & 0.597 $_{(15)}$ 
        & \multicolumn{1}{l}{ 0.482 $_{(13)}$ } & 0.472 $_{(14)}$ 
        & \multicolumn{1}{l}{ \rb{0.016} $_{(1)}$ } & 0.082 $_{(3)}$ 
        & \multicolumn{1}{l}{ 0.793 $_{(16)}$ } & 0.643 $_{(16)}$ 
        & \multicolumn{1}{l}{ 1.390 $_{(17)}$ } & 0.781 $_{(17)}$ & 12.7 \\

        PSA-GAN 
        & \multicolumn{1}{l}{ 0.623 $_{(13)}$ } & 0.546 $_{(13)}$ 
        & \multicolumn{1}{l}{ 0.537 $_{(15)}$ } & 0.488 $_{(15)}$ 
        & \multicolumn{1}{l}{ 0.018 $_{(7)}$ } & 0.087 $_{(7)}$ 
        & \multicolumn{1}{l}{ 0.535 $_{(14)}$ } & 0.533 $_{(14)}$ 
        & \multicolumn{1}{l}{ 1.220 $_{(15)}$ } & 0.578 $_{(16)}$ & 12.9 \\ 
        
        \midrule

        N-Hits 
        & \multicolumn{1}{l}{ 0.498 $_{(9)}$ } & 0.480 $_{(9)}$ 
        & \multicolumn{1}{l}{ 0.353 $_{(7)}$ } & 0.388 $_{(7)}$ 
        & \multicolumn{1}{l}{ 0.017 $_{(6)}$ } & 0.085 $_{(6)}$ 
        & \multicolumn{1}{l}{ 0.152 $_{(3)}$ } & \bu{0.245} $_{(2)}$ 
        & \multicolumn{1}{l}{ 0.323 $_{(5)}$ } & 0.335 $_{(5)}$ & 5.9 \\ 

        FiLM 
        & \multicolumn{1}{l}{ 0.426 $_{(6)}$ } & 0.436 $_{(5)}$ 
        & \multicolumn{1}{l}{ 0.347 $_{(5)}$ } & 0.374 $_{(4)}$ 
        & \multicolumn{1}{l}{ \rb{0.016} $_{(1)}$ } & \rb{0.079} $_{(1)}$ 
        & \multicolumn{1}{l}{ 0.210 $_{(7)}$ } & 0.320 $_{(7)}$ 
        & \multicolumn{1}{l}{ 0.327 $_{(6)}$ } & 0.336 $_{(6)}$ & 4.8 \\

        Depts 
        & \multicolumn{1}{l}{ 0.579 $_{(11)}$ } & 0.491 $_{(10)}$ 
        & \multicolumn{1}{l}{ 0.380 $_{(10)}$ } & 0.412 $_{(12)}$ 
        & \multicolumn{1}{l}{ 0.020 $_{(10)}$ } & 0.100 $_{(10)}$ 
        & \multicolumn{1}{l}{ 0.319 $_{(12)}$ } & 0.401 $_{(12)}$ 
        & \multicolumn{1}{l}{ 0.761 $_{(14)}$ } & 0.394 $_{(12)}$ & 11.3 \\ 

        NBeats 
        & \multicolumn{1}{l}{ 0.586$_{(12)}$ } & 0.521 $_{(12)}$ 
        & \multicolumn{1}{l}{ 0.391 $_{(11)}$ } & 0.409 $_{(11)}$ 
        & \multicolumn{1}{l}{ \rb{0.016} $_{(1)}$ } & \bu{0.081} $_{(2)}$ 
        & \multicolumn{1}{l}{ 0.269 $_{(10)}$ } & 0.370 $_{(10)}$ 
        & \multicolumn{1}{l}{ 1.344 $_{(16)}$ } & 0.420 $_{(13)}$ & 9.8 \\ 

        \midrule

        SCINet 
        & \multicolumn{1}{l}{ 0.465 $_{(7)}$ } & 0.463 $_{(8)}$ 
        & \multicolumn{1}{l}{ 0.359 $_{(8)}$ } & 0.389 $_{(8)}$ 
        & \multicolumn{1}{l}{ 0.036 $_{(12)}$ } & 0.137 $_{(13)}$ 
        & \multicolumn{1}{l}{ 0.171 $_{(5)}$ } & 0.280 $_{(5)}$ 
        & \multicolumn{1}{l}{ 0.329 $_{(7)}$ } & 0.344 $_{(7)}$ & 8 \\ 

        NLinear 
        & \multicolumn{1}{l}{ 0.410 $_{(3)}$ } & \bu{0.418} $_{(2)}$ 
        & \multicolumn{1}{l}{ 0.349 $_{(6)}$ } & 0.375 $_{(5)}$ 
        & \multicolumn{1}{l}{ 0.019 $_{(9)}$ } & 0.091 $_{(8)}$ 
        & \multicolumn{1}{l}{ \rb{0.147} $_{(1)}$ } & \rb{0.239} $_{(1)}$ 
        & \multicolumn{1}{l}{ 0.313 $_{(4)}$ } & 0.328 $_{(3)}$ & 4.2 \\ 

        DLinear 
        & \multicolumn{1}{l}{ 0.415 $_{(5)}$ } & 0.442 $_{(7)}$ 
        & \multicolumn{1}{l}{ 0.345 $_{(4)}$ } & 0.378 $_{(6)}$ 
        & \multicolumn{1}{l}{ 0.022 $_{(11)}$ } & 0.102 $_{(11)}$ 
        & \multicolumn{1}{l}{ 0.215 $_{(8)}$ } & 0.336 $_{(8)}$ 
        & \multicolumn{1}{l}{ 0.488 $_{(12)}$ } & 0.444 $_{(14)}$ & 8.6 \\ 

        LSTMa 
        & \multicolumn{1}{l}{ 1.149 $_{(17)}$ } & 0.782 $_{(17)}$ 
        & \multicolumn{1}{l}{ 1.030 $_{(17)}$ } & 0.699 $_{(17)}$ 
        & \multicolumn{1}{l}{ 0.403 $_{(17)}$ } & 0.534 $_{(17)}$ 
        & \multicolumn{1}{l}{ 0.414 $_{(13)}$ } & 0.444 $_{(13)}$ 
        & \multicolumn{1}{l}{ 0.662 $_{(13)}$ } & 0.501 $_{(15)}$ & 15.6 \\ 

        \bottomrule[1.2pt]
    \end{tabular}
    }

\end{table}

\begin{table}[!ht]
\centering
\caption{The inference time (in ms) of various time series generative models with different prediction lengths on the univariate ETTh1 dataset. The results of all baselines are taken from mr-Diff\cite{shen2024multi}.}
\vspace{-5pt}
\label{tab:inferencetime}
\resizebox{0.5\columnwidth}{!}{
\begin{tabular}{l|c c c c}
\toprule
\multicolumn{1}{c|}{} & 
\multicolumn{4}{c}{inference time (ms)}  \\
  & H = 96  & H = 192  & H = 336  & H = 720  \\

\cmidrule(lr){1-2} 
\cmidrule(lr){3-3}
\cmidrule(lr){4-4}
\cmidrule(lr){5-5}

TARFVAE  (\# of samples = 1)   & \textbf{2.9} & \textbf{3.0} & \textbf{2.8} & \textbf{3.0} \\
TARFVAE  (\# of samples = 50)  &  3.2 & 3.1 & 3.1 & 3.2 \\
TARFVAE  (\# of samples = 200)  &  3.2 & 3.2 & 3.1 & 3.2 \\
\midrule
mr-Diff(S=2)  & 8.3 & 9.8 & 11.9 & 21.6  \\ 
TimeDiff & 16.2 & 17.6 & 26.5 & 34.6  \\
TimeGrad  & 870.2 & 1854.5 & 3119.7  & 6724.1 \\
CSDI  & 90.4 & 142.8 & 398.9 & 513.1 \\
SSSD & 418.6 & 645.3 & 1054.2 & 2516.9 \\
\bottomrule
\end{tabular}}
\end{table}

 To ensure a systematic comparison of CRPS performance, we further benchmark our model against the baselines in ProbTS\cite{zhang2024probts} under an identical CRPS protocol. The results in \textbf{Table \ref{tab:crpscomparison}} show our TARFVAE still outperforms all probabilistic and point competitors in CRPS except for ranking 2nd in one case, demonstrating the effectiveness of our generative framework. To the best of our knowledge, this work represents the first demonstration of a generative model achieving comprehensive long-term forecasting across diverse datasets (with the number of channels ranging from 7 in ETT to 321 in Electricity) while simultaneously providing state-of-the-art deterministic (MSE, MAE) and probabilistic (CRPS) evaluation metrics.

\begin{table}[htbp] 
    \centering
    \caption{CRPS comparison following ProbTS\cite{zhang2024probts} protocol: TARFVAE vs. best baselines. The results of all baselines are taken from ProbTS\cite{zhang2024probts}.}
    \vspace{-5pt}
   \label{tab:crpscomparison}
   \resizebox{0.5\columnwidth}{!}{
\begin{tabular}{lcccc}
\toprule
\multirow{2}{*}{Dataset} & \multirow{2}{*}{Pred Len} & \multicolumn{2}{c}{CRPS} \\
\cmidrule(lr){3-4}
 & & TARFVAE & Other Top Performance \\
\midrule
\multirow{4}{*}{ETTm1} 
 & 96 & \textbf{0.222} & 0.236 (CSDI) \\
 & 192 & \textbf{0.244} & 0.291 (CSDI) \\
 & 336 & \textbf{0.261} & 0.322 (CSDI) \\
 & 720 & \textbf{0.291} & 0.353 (PatchTST) \\
\midrule

\multirow{4}{*}{ETTm2}
 & 96 & \textbf{0.110} & 0.115 (CSDI) \\
 & 192 & \textbf{0.131} & 0.147 (CSDI) \\
 & 336 & \textbf{0.150} & 0.176 (PatchTST) \\
 & 720 & \textbf{0.171} & 0.205 (PatchTST) \\
\midrule

\multirow{4}{*}{ETTh1}
 & 96 & \textbf{0.254} & 0.321 (iTransformer) \\
 & 192 & \textbf{0.278} & 0.359 (iTransformer \& PatchTST) \\
 & 336 & \textbf{0.301} & 0.384 (PatchTST) \\
 & 720 & \textbf{0.318} & 0.397 (PatchTST) \\
\midrule

\multirow{4}{*}{ETTh2}
 & 96 & \textbf{0.150} & 0.164 (CSDI) \\
 & 192 & \textbf{0.173} & 0.201 (PatchTST) \\
 & 336 & \textbf{0.185} & 0.240 (PatchTST) \\
 & 720 & \textbf{0.190} & 0.252 (PatchTST) \\
\midrule

\multirow{4}{*}{Electricity}
 & 96 & \textbf{0.068} & 0.086 (PatchTST) \\
 & 192 & \textbf{0.077} & 0.092 (PatchTST) \\
 & 336 & \textbf{0.081} & 0.099 (GRU NVP) \\
 & 720 & \textbf{0.087} & 0.108 (TimeGrad) \\
\midrule

\multirow{4}{*}{Weather}
 & 96 & \textbf{0.065} & 0.068 (CSDI) \\
 & 192 & 0.070 & \textbf{0.068} (CSDI) \\
 & 336 & \textbf{0.074} & 0.083 (CSDI) \\
 & 720 & \textbf{0.080} & 0.087 (CSDI) \\
\midrule

\multirow{4}{*}{Exchange}
 & 96 & \textbf{0.019} & 0.023 (PatchTST) \\
 & 192 & \textbf{0.028} & 0.034 (PatchTST) \\
 & 336 & \textbf{0.041} & 0.048 (iTransformer \& PatchTST \& DLinear) \\
 & 720 & \textbf{0.068} & 0.072 (PatchTST) \\
\bottomrule
\end{tabular}
}
\end{table}

\subsection{Ablation study}
\label{sec:ablation}
In this part, we conduct experiments using the exact same configuration as the main experiments, modifying only the targeted settings for analysis to ensure the consistency and reliability of the conclusions.

\begin{figure*}[!htbp]
  \centering
  \includegraphics[width=0.95\textwidth]{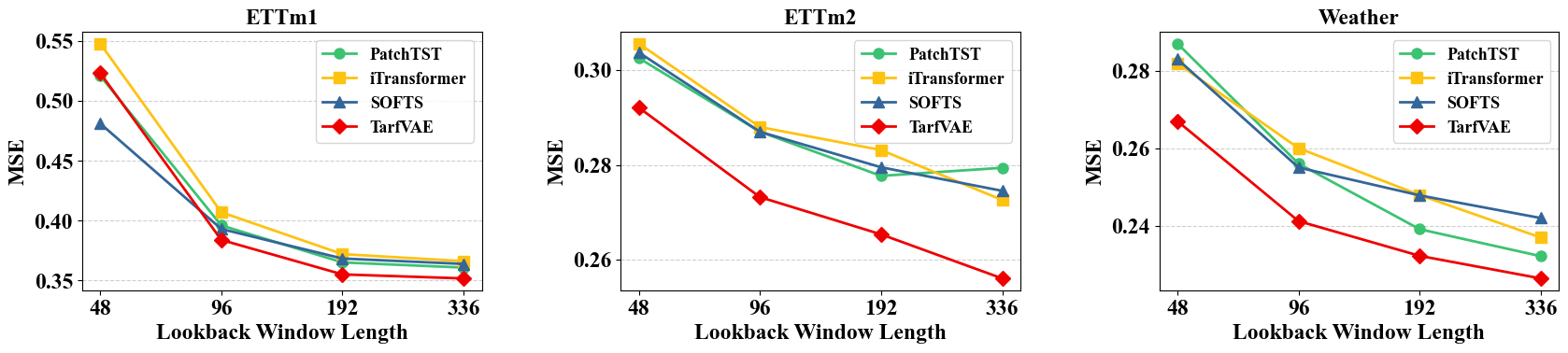}
  \caption{Influence of lookback window length $L$.}
  \label{fig:lbwmse}
\end{figure*}

\textbf{Influence of lookback windows}.
\textbf{Figure \ref{fig:lbwmse}} demonstrates the performance of TARFVAE with varying lookback window lengths. Overall, TARFVAE consistently achieves superior performance across all evaluated lookback windows, with its accuracy improving as the length gets longer. Notably, on the ETTm2 dataset, TARFVAE exhibits sustained performance enhancement with increasing window lengths, while competing methods show clear saturation effects.

\begin{table}[!ht]
\caption{Ablation study on various datasets. The results are averaged from forecast horizons $H \in \{ 96, 192, 336, 720 \}$ for all datasets with a fixed lookback window length $L = 96$.}
\vspace{-5pt}
\label{tab:ablation}
\resizebox{0.95\columnwidth}{!}{
\begin{tabular}{l|c|c|c|c|c|c|c|c|c|c|c|c|c|c|c|c||c|c}
\toprule
\multicolumn{1}{c|}{Datasets} & 
\multicolumn{2}{c|}{ETTh1} & 
\multicolumn{2}{c|}{ETTh2} & 
\multicolumn{2}{c|}{ETTm1} & 
\multicolumn{2}{c|}{ETTm2} & 
\multicolumn{2}{c|}{Exchange} & 
\multicolumn{2}{c|}{Electricity} & 
\multicolumn{2}{c|}{Solar} & 
\multicolumn{2}{c||}{Weather} & 
\multicolumn{2}{c}{\multirow{1}{*}{AVG}} \\

\midrule

\multicolumn{1}{c|}{Metric} & 
MSE & MAE & 
MSE & MAE & 
MSE & MAE & 
MSE & MAE & 
MSE & MAE & 
MSE & MAE & 
MSE & MAE & 
MSE & MAE &
MSE & MAE \\

\midrule

TARFVAE & 
 \rb{0.432}   &  \rb{0.434}  & \rb{0.356}  &   \rb{0.388} &  \rb{0.384}   &  \rb{0.394}  & \rb{0.273}  &   \rb{0.318}  & \rb{0.347}  &   \rb{0.307} & \rb{0.170}   &  \rb{0.266}  & \rb{0.233}  &   \rb{0.297} &  \rb{0.241}   &  \rb{0.268}  & \rb{0.301}  &   \rb{0.333} \\

w/o TARFLOW & 
\bu{0.441}  &   \bu{0.441} &  \bu{0.379}  &   \bu{0.404} &  0.396  &   \bu{0.402} &  \bu{0.279}  &   \bu{0.324}  &  0.420  &   0.323 & \bu{0.180}  &   \bu{0.278}&  0.316  &   0.300 &  \bu{0.243}  &   \bu{0.270}&  $\bu{0.332}^{(-10.30\%)}$  &   $\bu{0.343}^{(-3.00\%)}$ \\

w/o VAE\&TARFLOW   & 
0.466   &  0.457  &  0.386  &  0.410  &  \bu{0.394}  &  0.407  &  0.306  &  0.342 &  \bu{0.409}   &  \bu{0.322} & 0.186   &  0.283 &  \bu{0.275}   &  \bu{0.298} &  0.249   &  0.276 &  $0.334^{(-10.96\%)}$  &  $0.349^{(-4.80\%)}$  \\

\bottomrule
\end{tabular}}
\end{table}

\textbf{Architecture ablations}. As evidenced in \textbf{Table \ref{tab:ablation}}, removing the TARFLOW component and further eliminating VAE lead to a sustained deterioration. These findings collectively demonstrate that the synergistic integration of TARFLOW with VAE critically enables TARFVAE's superior modeling capacity.

\textbf{Influence of inference sample size}. Upon completing model training, we evaluate TARFVAE's inference performance under varying sample sizes, as shown in \textbf{Table \ref{tab:crps}}. As the sample size increases from 20 to 200, the reduced sampling bias consistently improves the MSE, MAE and CRPS. The absence of significant metric fluctuations across different sample sizes confirms the stability of TARFVAE's performance. These collective findings substantiate that TARFVAE achieves both reliable uncertainty quantification and deterministic modeling with strong operational stability.

\section{Conclusion}
In this work we present TARFVAE, an efficient one-step generative framework for time series forecasting. It integrates TARFLOW with VAE architecture to enhance posterior estimation, enabling the learning of a more expressive latent space which can lead to efficient modeling of the targeted time series distribution. Implemented on an simple MLP foundation in our extensive experiments, TARFVAE achieves superior or comparable‌ performance to recent state-of-the-art deterministic and generative models across widely-used real-world benchmarks. To our knowledge, this is the first time a generative time series forecasting model delivers comprehensive long-term predictions consistent with deterministic methods on rich datasets, encompassing both deterministic (MSE, MAE) and probabilistic (CRPS) forecasting metrics. Meanwhile, TARFVAE's one-step generation reduces inference time by over $85\%$ compared to other state-of-the-art generative models in long-term forecasting. Other implementations could be tried to achieve better performance in the future.

\bibliographystyle{unsrtnat}
\bibliography{neurips_2025}

\clearpage

\section*{NeurIPS Paper Checklist}

\begin{enumerate}

\item {\bf Claims}
    \item[] Question: Do the main claims made in the abstract and introduction accurately reflect the paper's contributions and scope?
    \item[] Answer: \answerYes{} 
    \item[] Justification: The abstract and introduction clearly help readers to understand our contributions and scope of TARFVAE, a novel model combining TARFlow and conditional VAE designed for long-term forecasting and uncertainty quantification.
    \item[] Guidelines:
    \begin{itemize}
        \item The answer NA means that the abstract and introduction do not include the claims made in the paper.
        \item The abstract and/or introduction should clearly state the claims made, including the contributions made in the paper and important assumptions and limitations. A No or NA answer to this question will not be perceived well by the reviewers. 
        \item The claims made should match theoretical and experimental results, and reflect how much the results can be expected to generalize to other settings. 
        \item It is fine to include aspirational goals as motivation as long as it is clear that these goals are not attained by the paper. 
    \end{itemize}

\item {\bf Limitations}
    \item[] Question: Does the paper discuss the limitations of the work performed by the authors?
    \item[] Answer: \answerYes{} 
    \item[] Justification: We have discussed the limitations of our work in Appendix \ref{apd:limitaion}.
    \item[] Guidelines:
    \begin{itemize}
        \item The answer NA means that the paper has no limitation while the answer No means that the paper has limitations, but those are not discussed in the paper. 
        \item The authors are encouraged to create a separate "Limitations" section in their paper.
        \item The paper should point out any strong assumptions and how robust the results are to violations of these assumptions (e.g., independence assumptions, noiseless settings, model well-specification, asymptotic approximations only holding locally). The authors should reflect on how these assumptions might be violated in practice and what the implications would be.
        \item The authors should reflect on the scope of the claims made, e.g., if the approach was only tested on a few datasets or with a few runs. In general, empirical results often depend on implicit assumptions, which should be articulated.
        \item The authors should reflect on the factors that influence the performance of the approach. For example, a facial recognition algorithm may perform poorly when image resolution is low or images are taken in low lighting. Or a speech-to-text system might not be used reliably to provide closed captions for online lectures because it fails to handle technical jargon.
        \item The authors should discuss the computational efficiency of the proposed algorithms and how they scale with dataset size.
        \item If applicable, the authors should discuss possible limitations of their approach to address problems of privacy and fairness.
        \item While the authors might fear that complete honesty about limitations might be used by reviewers as grounds for rejection, a worse outcome might be that reviewers discover limitations that aren't acknowledged in the paper. The authors should use their best judgment and recognize that individual actions in favor of transparency play an important role in developing norms that preserve the integrity of the community. Reviewers will be specifically instructed to not penalize honesty concerning limitations.
    \end{itemize}

\item {\bf Theory assumptions and proofs}
    \item[] Question: For each theoretical result, does the paper provide the full set of assumptions and a complete (and correct) proof?
    \item[] Answer: \answerYes{}{} 
    \item[] Justification: We provide numerous formulas to support the complete derivation of the theory in Section \ref{sec:preliminaries} and Section \ref{sec:tarflow}.
    \item[] Guidelines:
    \begin{itemize}
        \item The answer NA means that the paper does not include theoretical results. 
        \item All the theorems, formulas, and proofs in the paper should be numbered and cross-referenced.
        \item All assumptions should be clearly stated or referenced in the statement of any theorems.
        \item The proofs can either appear in the main paper or the supplemental material, but if they appear in the supplemental material, the authors are encouraged to provide a short proof sketch to provide intuition. 
        \item Inversely, any informal proof provided in the core of the paper should be complemented by formal proofs provided in appendix or supplemental material.
        \item Theorems and Lemmas that the proof relies upon should be properly referenced. 
    \end{itemize}

    \item {\bf Experimental result reproducibility}
    \item[] Question: Does the paper fully disclose all the information needed to reproduce the main experimental results of the paper to the extent that it affects the main claims and/or conclusions of the paper (regardless of whether the code and data are provided or not)?
    \item[] Answer: \answerYes{} 
    \item[] Justification: We provide full information needed to reproduce the main experimental results in Section \ref{sec:setup}.
    \item[] Guidelines: 
    \begin{itemize}
        \item The answer NA means that the paper does not include experiments.
        \item If the paper includes experiments, a No answer to this question will not be perceived well by the reviewers: Making the paper reproducible is important, regardless of whether the code and data are provided or not.
        \item If the contribution is a dataset and/or model, the authors should describe the steps taken to make their results reproducible or verifiable. 
        \item Depending on the contribution, reproducibility can be accomplished in various ways. For example, if the contribution is a novel architecture, describing the architecture fully might suffice, or if the contribution is a specific model and empirical evaluation, it may be necessary to either make it possible for others to replicate the model with the same dataset, or provide access to the model. In general. releasing code and data is often one good way to accomplish this, but reproducibility can also be provided via detailed instructions for how to replicate the results, access to a hosted model (e.g., in the case of a large language model), releasing of a model checkpoint, or other means that are appropriate to the research performed.
        \item While NeurIPS does not require releasing code, the conference does require all submissions to provide some reasonable avenue for reproducibility, which may depend on the nature of the contribution. For example
        \begin{enumerate}
            \item If the contribution is primarily a new algorithm, the paper should make it clear how to reproduce that algorithm.
            \item If the contribution is primarily a new model architecture, the paper should describe the architecture clearly and fully.
            \item If the contribution is a new model (e.g., a large language model), then there should either be a way to access this model for reproducing the results or a way to reproduce the model (e.g., with an open-source dataset or instructions for how to construct the dataset).
            \item We recognize that reproducibility may be tricky in some cases, in which case authors are welcome to describe the particular way they provide for reproducibility. In the case of closed-source models, it may be that access to the model is limited in some way (e.g., to registered users), but it should be possible for other researchers to have some path to reproducing or verifying the results.
        \end{enumerate}
    \end{itemize}

\item {\bf Open access to data and code}
    \item[] Question: Does the paper provide open access to the data and code, with sufficient instructions to faithfully reproduce the main experimental results, as described in supplemental material?
    \item[] Answer: \answerYes{} 
    \item[] Justification: All experiments are based on open-source data, and our reproducible code will be made available through an anonymous GitHub repository. The code will be made public once the paper is accepted.
    \item[] Guidelines:
    \begin{itemize}
        \item The answer NA means that paper does not include experiments requiring code.
        \item Please see the NeurIPS code and data submission guidelines (\url{https://nips.cc/public/guides/CodeSubmissionPolicy}) for more details.
        \item While we encourage the release of code and data, we understand that this might not be possible, so “No” is an acceptable answer. Papers cannot be rejected simply for not including code, unless this is central to the contribution (e.g., for a new open-source benchmark).
        \item The instructions should contain the exact command and environment needed to run to reproduce the results. See the NeurIPS code and data submission guidelines (\url{https://nips.cc/public/guides/CodeSubmissionPolicy}) for more details.
        \item The authors should provide instructions on data access and preparation, including how to access the raw data, preprocessed data, intermediate data, and generated data, etc.
        \item The authors should provide scripts to reproduce all experimental results for the new proposed method and baselines. If only a subset of experiments are reproducible, they should state which ones are omitted from the script and why.
        \item At submission time, to preserve anonymity, the authors should release anonymized versions (if applicable).
        \item Providing as much information as possible in supplemental material (appended to the paper) is recommended, but including URLs to data and code is permitted.
    \end{itemize}

\item {\bf Experimental setting/details}
    \item[] Question: Does the paper specify all the training and test details (e.g., data splits, hyperparameters, how they were chosen, type of optimizer, etc.) necessary to understand the results?
    \item[] Answer: \answerYes{} 
    \item[] Justification: We provide a detailed description of the experimental setup in Section \ref{sec:setup}.
    \item[] Guidelines:
    \begin{itemize}
        \item The answer NA means that the paper does not include experiments.
        \item The experimental setting should be presented in the core of the paper to a level of detail that is necessary to appreciate the results and make sense of them.
        \item The full details can be provided either with the code, in appendix, or as supplemental material.
    \end{itemize}

\item {\bf Experiment statistical significance}
    \item[] Question: Does the paper report error bars suitably and correctly defined or other appropriate information about the statistical significance of the experiments?
    \item[] Answer: \answerYes{} 
    \item[] Justification: In section \ref{sec:ablation}, we conduct ablation study using the exact same configuration as the main experiments, modifying only the targeted settings for analysis to ensure the consistency and reliability of the conclusions.
    \item[] Guidelines:
    \begin{itemize}
        \item The answer NA means that the paper does not include experiments.
        \item The authors should answer "Yes" if the results are accompanied by error bars, confidence intervals, or statistical significance tests, at least for the experiments that support the main claims of the paper.
        \item The factors of variability that the error bars are capturing should be clearly stated (for example, train/test split, initialization, random drawing of some parameter, or overall run with given experimental conditions).
        \item The method for calculating the error bars should be explained (closed form formula, call to a library function, bootstrap, etc.)
        \item The assumptions made should be given (e.g., Normally distributed errors).
        \item It should be clear whether the error bar is the standard deviation or the standard error of the mean.
        \item It is OK to report 1-sigma error bars, but one should state it. The authors should preferably report a 2-sigma error bar than state that they have a 96\% CI, if the hypothesis of Normality of errors is not verified.
        \item For asymmetric distributions, the authors should be careful not to show in tables or figures symmetric error bars that would yield results that are out of range (e.g. negative error rates).
        \item If error bars are reported in tables or plots, The authors should explain in the text how they were calculated and reference the corresponding figures or tables in the text.
    \end{itemize}

\item {\bf Experiments compute resources}
    \item[] Question: For each experiment, does the paper provide sufficient information on the computer resources (type of compute workers, memory, time of execution) needed to reproduce the experiments?
    \item[] Answer: \answerYes{} 
    \item[] Justification: Section \ref{sec:setup} specifies the runtime environment and computational resources employed.
    \item[] Guidelines:
    \begin{itemize}
        \item The answer NA means that the paper does not include experiments.
        \item The paper should indicate the type of compute workers CPU or GPU, internal cluster, or cloud provider, including relevant memory and storage.
        \item The paper should provide the amount of compute required for each of the individual experimental runs as well as estimate the total compute. 
        \item The paper should disclose whether the full research project required more compute than the experiments reported in the paper (e.g., preliminary or failed experiments that didn't make it into the paper). 
    \end{itemize}
    
\item {\bf Code of ethics}
    \item[] Question: Does the research conducted in the paper conform, in every respect, with the NeurIPS Code of Ethics \url{https://neurips.cc/public/EthicsGuidelines}?
    \item[] Answer: \answerYes{} 
    \item[] Justification: We conform with the NeurIPS Code of Ethics in every respect.
    \item[] Guidelines:
    \begin{itemize}
        \item The answer NA means that the authors have not reviewed the NeurIPS Code of Ethics.
        \item If the authors answer No, they should explain the special circumstances that require a deviation from the Code of Ethics.
        \item The authors should make sure to preserve anonymity (e.g., if there is a special consideration due to laws or regulations in their jurisdiction).
    \end{itemize}

\item {\bf Broader impacts}
    \item[] Question: Does the paper discuss both potential positive societal impacts and negative societal impacts of the work performed?
    \item[] Answer: \answerYes{} 
    \item[] Justification: We have discussed both potential positive societal impacts and negative societal impacts of our work in Appendix \ref{apd:impacts}.
    \item[] Guidelines:
    \begin{itemize}
        \item The answer NA means that there is no societal impact of the work performed.
        \item If the authors answer NA or No, they should explain why their work has no societal impact or why the paper does not address societal impact.
        \item Examples of negative societal impacts include potential malicious or unintended uses (e.g., disinformation, generating fake profiles, surveillance), fairness considerations (e.g., deployment of technologies that could make decisions that unfairly impact specific groups), privacy considerations, and security considerations.
        \item The conference expects that many papers will be foundational research and not tied to particular applications, let alone deployments. However, if there is a direct path to any negative applications, the authors should point it out. For example, it is legitimate to point out that an improvement in the quality of generative models could be used to generate deepfakes for disinformation. On the other hand, it is not needed to point out that a generic algorithm for optimizing neural networks could enable people to train models that generate Deepfakes faster.
        \item The authors should consider possible harms that could arise when the technology is being used as intended and functioning correctly, harms that could arise when the technology is being used as intended but gives incorrect results, and harms following from (intentional or unintentional) misuse of the technology.
        \item If there are negative societal impacts, the authors could also discuss possible mitigation strategies (e.g., gated release of models, providing defenses in addition to attacks, mechanisms for monitoring misuse, mechanisms to monitor how a system learns from feedback over time, improving the efficiency and accessibility of ML).
    \end{itemize}
    
\item {\bf Safeguards}
    \item[] Question: Does the paper describe safeguards that have been put in place for responsible release of data or models that have a high risk for misuse (e.g., pretrained language models, image generators, or scraped datasets)?
    \item[] Answer: \answerNA{} 
    \item[] Justification: The paper poses no such risks.
    \item[] Guidelines:
    \begin{itemize}
        \item The answer NA means that the paper poses no such risks.
        \item Released models that have a high risk for misuse or dual-use should be released with necessary safeguards to allow for controlled use of the model, for example by requiring that users adhere to usage guidelines or restrictions to access the model or implementing safety filters. 
        \item Datasets that have been scraped from the Internet could pose safety risks. The authors should describe how they avoided releasing unsafe images.
        \item We recognize that providing effective safeguards is challenging, and many papers do not require this, but we encourage authors to take this into account and make a best faith effort.
    \end{itemize}

\item {\bf Licenses for existing assets}
    \item[] Question: Are the creators or original owners of assets (e.g., code, data, models), used in the paper, properly credited and are the license and terms of use explicitly mentioned and properly respected?
    \item[] Answer: \answerYes{} 
    \item[] Justification: We cite the original paper for all the baselines properly.
    \item[] Guidelines:
    \begin{itemize}
        \item The answer NA means that the paper does not use existing assets.
        \item The authors should cite the original paper that produced the code package or dataset.
        \item The authors should state which version of the asset is used and, if possible, include a URL.
        \item The name of the license (e.g., CC-BY 4.0) should be included for each asset.
        \item For scraped data from a particular source (e.g., website), the copyright and terms of service of that source should be provided.
        \item If assets are released, the license, copyright information, and terms of use in the package should be provided. For popular datasets, \url{paperswithcode.com/datasets} has curated licenses for some datasets. Their licensing guide can help determine the license of a dataset.
        \item For existing datasets that are re-packaged, both the original license and the license of the derived asset (if it has changed) should be provided.
        \item If this information is not available online, the authors are encouraged to reach out to the asset's creators.
    \end{itemize}

\item {\bf New assets}
    \item[] Question: Are new assets introduced in the paper well documented and is the documentation provided alongside the assets?
    \item[] Answer: \answerNA{} 
    \item[] Justification: The paper does not release new assets.
    \item[] Guidelines:
    \begin{itemize}
        \item The answer NA means that the paper does not release new assets.
        \item Researchers should communicate the details of the dataset/code/model as part of their submissions via structured templates. This includes details about training, license, limitations, etc. 
        \item The paper should discuss whether and how consent was obtained from people whose asset is used.
        \item At submission time, remember to anonymize your assets (if applicable). You can either create an anonymized URL or include an anonymized zip file.
    \end{itemize}

\item {\bf Crowdsourcing and research with human subjects}
    \item[] Question: For crowdsourcing experiments and research with human subjects, does the paper include the full text of instructions given to participants and screenshots, if applicable, as well as details about compensation (if any)? 
    \item[] Answer: \answerNA{} 
    \item[] Justification: The paper does not involve crowdsourcing nor research with human subjects.
    \item[] Guidelines:
    \begin{itemize}
        \item The answer NA means that the paper does not involve crowdsourcing nor research with human subjects.
        \item Including this information in the supplemental material is fine, but if the main contribution of the paper involves human subjects, then as much detail as possible should be included in the main paper. 
        \item According to the NeurIPS Code of Ethics, workers involved in data collection, curation, or other labor should be paid at least the minimum wage in the country of the data collector. 
    \end{itemize}

\item {\bf Institutional review board (IRB) approvals or equivalent for research with human subjects}
    \item[] Question: Does the paper describe potential risks incurred by study participants, whether such risks were disclosed to the subjects, and whether Institutional Review Board (IRB) approvals (or an equivalent approval/review based on the requirements of your country or institution) were obtained?
    \item[] Answer: \answerNA{} 
    \item[] Justification: The paper does not involve crowdsourcing nor research with human subjects.
    \item[] Guidelines:
    \begin{itemize}
        \item The answer NA means that the paper does not involve crowdsourcing nor research with human subjects.
        \item Depending on the country in which research is conducted, IRB approval (or equivalent) may be required for any human subjects research. If you obtained IRB approval, you should clearly state this in the paper. 
        \item We recognize that the procedures for this may vary significantly between institutions and locations, and we expect authors to adhere to the NeurIPS Code of Ethics and the guidelines for their institution. 
        \item For initial submissions, do not include any information that would break anonymity (if applicable), such as the institution conducting the review.
    \end{itemize}

\item {\bf Declaration of LLM usage}
    \item[] Question: Does the paper describe the usage of LLMs if it is an important, original, or non-standard component of the core methods in this research? Note that if the LLM is used only for writing, editing, or formatting purposes and does not impact the core methodology, scientific rigorousness, or originality of the research, declaration is not required.
    \item[] Answer: \answerNA{} 
    \item[] Justification: The paper does not involve LLMs as any important, original, or non-standard components.
    \item[] Guidelines:
    \begin{itemize}
        \item The answer NA means that the core method development in this research does not involve LLMs as any important, original, or non-standard components.
        \item Please refer to our LLM policy (\url{https://neurips.cc/Conferences/2025/LLM}) for what should or should not be described.
    \end{itemize}

\end{enumerate}

\clearpage

\appendix

\section{Calculation of CRPS}
\label{apd:crpsmethod}
After the inference of sufficient samples, we employ an efficient quantile-based approach to calculate CRPS. This aims to mitigate computation costs in estimating CRPS with all samples. Given certain quantiles $[q_1,q_2,...,q_n]$, we
first sort these samples and extract corresponding quantile values $[Q(q_1),Q(q_2),...,Q(q_n)]$, then compute CRPS using the following formulation:
\begin{align}
    CRPS=\int_0^1 2\cdot(Q(q)-y)\cdot(I(y\le Q(q))-q)\mathrm{d}q
    \label{eq:quantilecrps}
\end{align}
We adopt a constant extrapolation, that is $Q(q)=Q(q_1), \forall q<q_1$ and $Q(q)=Q(q_n), \forall q>q_n$. Thus, (\ref{eq:quantilecrps}) can be further formulated as
\begin{align}
    CRPS &=\int_0^{q_1} 2\cdot(Q(q_1)-y)\cdot(I(y\le Q(q_1))-q)\mathrm{d}q \label{eq:crpslow}\\
    &\quad + \int_{q_1}^{q_n} 2\cdot(Q(q)-y)\cdot(I(y\le Q(q))-q)\mathrm{d}q \label{eq:crpsmid}\\
    &\quad + \int_{q_n}^{1} 2\cdot(Q(q_n)-y)\cdot(I(y\le Q(q_n))-q)\mathrm{d}q. \label{eq:crpshigh}
\end{align}
As for (\ref{eq:crpslow}), it equals to
\begin{align}
\begin{cases}
2\cdot(Q(q_1)-y)\cdot(q_1-0.5q_1^2), &y\le Q(q_1);\\
2\cdot(Q(q_1)-y)\cdot(-0.5q_1^2), &y> Q(q_1).
\end{cases}
\end{align}
As for (\ref{eq:crpshigh}), it equals to
\begin{align}
\begin{cases}
2\cdot(Q(q_n)-y)\cdot(0.5(1-q_n)^2), &y\le Q(q_n);\\
2\cdot(Q(q_n)-y)\cdot(-0.5(1-q_n^2)), &y> Q(q_n).
\end{cases}
\end{align}
As for (\ref{eq:crpsmid}), it is discretized and approximated as
\begin{align}
2\cdot\sum_{i=1}^{n}(Q(q_i)-y)\cdot(I(y\le Q(q_i))-q_i)\cdot \Delta q.
\end{align}
The inplementation of this calculation process can be found in our code. 

\section{Limitations and Future Works}
\label{apd:limitaion}
We analyze the limitations of our work and briefly discuss several directions for future research. In our work, we employ identical MLP blocks for the encoder, decoder, and prior modules. This configuration might limit the forecasting performance of the proposed TARFVAE model. For instance, since the encoder, decoder, and prior typically perform different functions, we may flexibly design distinct architectures for these three components. Moreover, MLP blocks may not be sufficiently powerful to effectively extract time series patterns from the input data, potentially hindering TARFVAE's ability to learn complex time series patterns. In the future, exploring advanced temporal processing modules or deep learning models within the TARFVAE framework could be beneficial.

\section{Societal Impacts}
\label{apd:impacts}
Extensive experiments on real-world datasets demonstrate the TARFVAE's potential capability to significantly benefit various application fields, spanning from finance to energy, by providing accurate and reliable forecasts. However, there are potential negative societal impacts to consider. As a model that often interacts with sensitive and proprietary data, particularly in sectors like healthcare and finance, TARFVAE could inadvertently contribute to privacy risks if the data it processes is not adequately protected. Moreover, Organizations might also become overly reliant on automated forecasts, neglecting valuable insights from human experts, which can affect decision quality. To prevent the aforementioned negative societal impacts, data integrity and security strategies should be well-designed to ensure the deployment of TARFVAE enhances its beneficial societal effects while reducing possible adverse impacts.

\end{document}